\documentclass[11pt]{article}

\usepackage[final]{acl}

\usepackage{times}
\usepackage{latexsym}

\usepackage[T1]{fontenc}

\usepackage[utf8]{inputenc}

\usepackage{microtype}

\usepackage{inconsolata}

\usepackage{graphicx}
\usepackage{hyperref}
\usepackage{url}
\usepackage{enumitem}
\usepackage{graphicx}
\newcommand{\hide}[1]{}
\usepackage{wrapfig} 
\usepackage{multirow}
\usepackage{inconsolata}
\usepackage{calligra}
\usepackage{helvet}
\usepackage{newtxtt}
\usepackage{amssymb}
\usepackage{amsxtra}
\usepackage{multirow}
\usepackage{diagbox}
\usepackage{longtable}
\usepackage{array}
\usepackage{stfloats}
\usepackage{graphicx}
\usepackage{multicol}
\usepackage{amsmath, bm}
\usepackage{booktabs} 
\usepackage{arydshln}
\usepackage{xspace}
\usepackage{enumitem}
\usepackage{threeparttable}
\usepackage{dsfont}
\usepackage[most]{tcolorbox}
\usepackage{xcolor}  
\usepackage{pgf-pie}

\usepackage{bbm}
\usepackage{makecell}
\usepackage[linesnumbered,ruled,vlined]{algorithm2e}
\usepackage{color} 
\usepackage{CJKutf8}

\newcommand{\vpara}[1]{\vspace{0.05in}\noindent \textbf{#1 }}

\title{GeoVLMath: Enhancing Geometry Reasoning in Vision-Language Models via Cross-Modal Reward for Auxiliary Line Creation}

 \author{Shasha Guo\textsuperscript{1}, 
 Liang Pang\textsuperscript{1}, Xi Wang\textsuperscript{2}, 
\textbf{Yanling Wang\textsuperscript{3}}, \textbf{Huawei Shen\textsuperscript{1}}, \textbf{Jing Zhang\textsuperscript{2}\thanks{       $\text{ }$ Corresponding Author\normalsize}} \\
  \textsuperscript{1}Institute of Computing Technology, Chinese Academy of Sciences, China\\
  \textsuperscript{2}Renmin University of China, China  \textsuperscript{3}Zhipu AI, China\\
  \{guoshasha, pangliang, shenhuawei\}@ict.ac.cn\\
  \{wangxi2022, zhang-jing\}@ruc.edu.cn, yanlingwang777@gmail.com
  }

\begin{document}
\maketitle
\begin{abstract}

Auxiliary lines are essential for solving complex geometric problems but remain challenging for large vision-language models (LVLMs). 
Recent attempts construct auxiliary lines via code-driven rendering, a strategy that relies on accurate and executable code generation to produce visual renderings of the auxiliary lines for subsequent reasoning. However, in complex solid geometry settings, such a strong dependence on precise specifications substantially restricts the robustness of this strategy. Alternatively, we turn to a simpler and more stable solution, representing auxiliary-line constructions as structured textual descriptions.
To bridge the gap between textual descriptions and spatial structure, we propose a reinforcement learning framework that enhances diagram-text alignment. 
The core is a cross-modal reward model that evaluates how well the generated auxiliary-line description matches the ground-truth auxiliary-line diagram. 
The reward signal drives a GRPO-based RL stage to  yield informative auxiliary-line descriptions for the reasoning.
To support the training and evaluation, we develop a scalable data pipeline and construct AuxSolidMath\footnote{The dataset is now publicly available at https://huggingface.co/datasets/shasha/AuxSolidMath}, a dataset of 3,018 real-exam geometry problems with paired diagrams and aligned textual fields. Based on this framework, we derive GeoVLMath, an LVLM for solving complex solid geometry.

\end{abstract}
\section{Introduction}
\label{sec:intro}

Geometric problems constitute an important category of mathematical tasks, characterized by intricate spatial structures and multi-step reasoning processes~\citep{ma2024geometric}. They are commonly divided into plane geometry and solid geometry. This study focuses on solid geometry, where reasoning over three-dimensional spatial relations is substantially more complex. Such problems rarely yield to direct application of standard theorems; instead, they often require the deliberate introduction of auxiliary lines\footnote{In this paper, we use \textit{auxiliary lines} broadly to include both additional lines and coordinate systems.} to reveal hidden geometric structure and enable further analysis.
These auxiliary lines are essential for anchoring visual diagrams to formal symbolic reasoning and for providing the intermediate steps for rigorous problem-solving.

\begin{figure}[!t]
\centering 
 \includegraphics[width=0.43\textwidth]{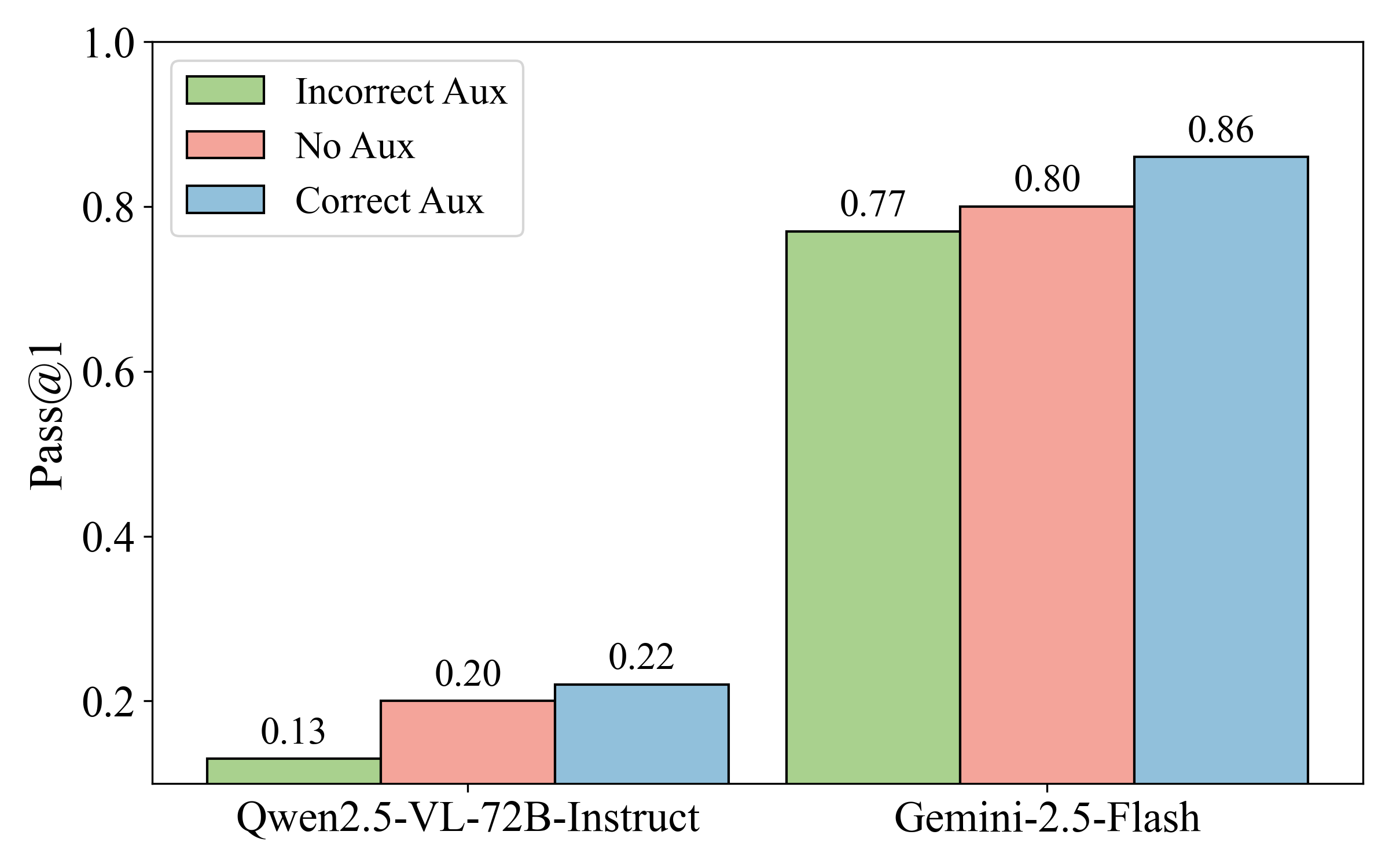}
\caption{Pass@1 results for Qwen2.5-VL-72B-Instruct and Gemini-2.5-Flash. ``Aux'' indicates the use of auxiliary-line descriptions.}
\vspace{-15pt}
\label{fig:pilot_study}
\end{figure}

To validate the above idea, we conduct a pilot study comprising three experimental settings: Incorrect Aux (with incorrect auxiliary lines), No Aux (without auxiliary lines), and Correct Aux (with correct auxiliary lines).
As shown in Figure~\ref{fig:pilot_study}, \textbf{the use of correct auxiliary lines achieves the highest accuracy, whereas incorrect auxiliary lines lead to the poorest performance.}
One possible explanation is that inaccurate auxiliary lines tend to misdirect reasoning and produce errors, while precise auxiliary lines uncover key spatial relationships, thereby enhancing solution accuracy.


Given that accurate auxiliary lines are crucial for the correctness of the solution, the key question is how to obtain them reliably. A seemingly straightforward solution is to explicitly construct them on the diagram. Indeed, several representative methods, such as Visual Sketchpad~\citep{Visual_Sketchpad}, V-Thinker~\citep{v-thinker}, and CodePlot-CoT~\citep{CodePlot-CoT}, follow a unified paradigm in which the model generates Python-based drawing code to render intermediate diagrams that are then fed back into the reasoning process. 
Despite its conceptual simplicity, this code-driven visual construction paradigm exhibits inherent limitations in solid geometry. It critically relies on precise coordinate information and highly accurate code generation, and becomes particularly fragile when auxiliary lines involve cross-plane relations or skew structures, where minor rendering inaccuracies can distort spatial constraints and mislead reasoning.
Our empirical results corroborate this limitation, \textbf{showing that explicitly rendered auxiliary-line diagrams underperform text-based auxiliary-line formulations} (See Table~\ref{tb:overall_evaluation2}). 

In light of these observations, we adopt a simpler and more stable formulation by representing auxiliary-line constructions as structured textual descriptions.
Our main idea is to design a \textbf{cross-modal reward model} that measures the consistency between a generated textual auxiliary-line description for the original diagram and a ground-truth auxiliary-line diagram. 
The reward is computed by jointly encoding the original diagram and the generated auxiliary-line description, and then comparing this pair with the ground-truth auxiliary-line diagram, providing geometry-aware supervision without requiring coordinate assumptions or image manipulation. 
Building on this reward signal, we train a policy model using Group Relative Policy Optimization (GRPO) to obtain geometry-consistent, generalizable constructions.
Training follows a two-stage paradigm inspired by recent progress in reinforcement learning (RL) for reasoning (e.g., DeepSeek-R1~\citep{deepseekr1}): supervised fine-tuning (SFT) for cold start, followed by GRPO-based RL to further elicit structured reasoning and strengthen diagram-text alignment. We instantiate the framework as GeoVLMath, a vision-language model tailored to auxiliary-line-based geometric reasoning. Through the cross-modal supervision, GeoVLMath achieves strong alignment between text and geometric structure, enabling faithful reasoning on complex diagrams.


To effectively train the above model, we require a high-quality dataset that captures both visual and symbolic aspects of real-world geometry problems. However, creating such a dataset is inherently challenging due to the need for automation, scalability, and semantic precision across diverse and noisy educational materials.
In response to these challenges, we develop a \textbf{robust and scalable data construction pipeline} that transforms raw high school exam papers into structured multimodal instances suitable for training LVLMs, comprising automated problem identification, automated deduplication and diagram extraction, structured data extraction, and manual verification.
While the pipeline is largely automated, this lightweight manual verification step is essential for maintaining data quality, particularly when handling complex symbolic expressions and diagrammatic content in real-world settings.
Based on this pipeline, we construct \textbf{AuxSolidMath}, a curated dataset of 3,018 solid geometry problems in a rich multimodal format, comprising the problem description, the final answer, the auxiliary-line description, the original diagram, and the auxiliary-line diagram. To our knowledge, AuxSolidMath is the first systematically constructed dataset explicitly tailored to auxiliary-line-based solid geometry reasoning. 

We empirically evaluate GeoVLMath against a broad range of LVLMs, including representative approaches that construct auxiliary lines via code-driven rendering. Despite its relatively modest parameter scale, GeoVLMath achieves highly competitive performance, consistently outperforming code-driven rendering methods and surpassing larger models such as Qwen2.5-VL-32B-Instruct~\citep{Qwen2.5vl} and GPT-4o~\citep{Gpt4o}.
These results indicate that supervision grounded in auxiliary-line constructions is more effective for improving geometric reasoning than explicit code-driven diagram rendering or simply scaling model size.
Furthermore, our evaluation protocol highlights the value of auxiliary-line-augmented datasets for revealing the limitations of LVLMs in visual reasoning.

\vpara{Contributions.} (1) \textbf{Cross-modal reward.} We introduce a geometry-aware scalar reward that directly evaluates diagram-text alignment, providing reliable and fine-grained supervision for reinforcement learning in auxiliary-line construction.
(2) \textbf{AuxSolidMath.} We introduce AuxSolidMath, a curated dataset of 3,018 solid geometry problems from real high-school exams, with aligned diagrams, auxiliary-line annotations, and answers, designed to support training and evaluation of auxiliary-line reasoning.
(3) \textbf{GeoVLMath.} We introduce GeoVLMath, demonstrating that RL with vision-based rewards effectively optimizes auxiliary-line construction, achieving superior performance to code-driven rendering approaches and competitive results against larger LVLMs.

\section{Methodology}
\label{sec:method}
\subsection{Problem Definition}
We study solid geometry problems whose solutions require the \emph{active construction of auxiliary lines}. Each instance is defined by a pair $\langle I, q \rangle$, where $I$ denotes the original diagram and $q$ is a question grounded in $I$. The goal is to generate a solution $y$ that includes the auxiliary lines $aux$, a sequence of reasoning steps, and the final answer $ans$.
The $aux$ is not present in $I$ and must be introduced during reasoning to expose implicit spatial relations and convert them into explicit geometric constraints.

\begin{figure*}[ht]
\centering 
\includegraphics[width=0.85\textwidth]{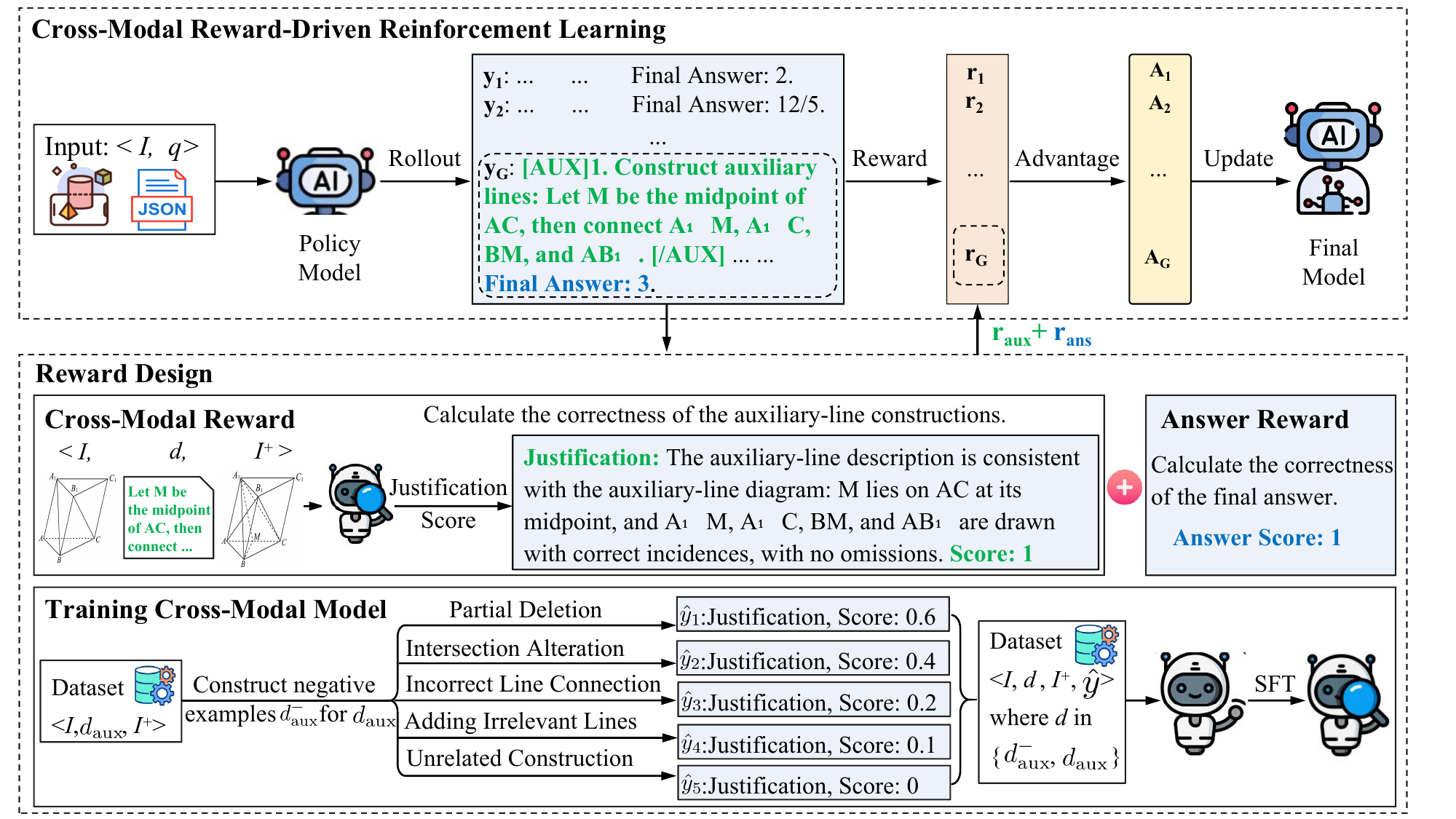}
\caption{Overview of the cross-modal reward-driven RL. We first fine-tune a cross-modal reward model to evaluate the correctness of auxiliary-line constructions. During the RL phase, the reward model’s consistency score is combined with the answer accuracy reward to produce a composite signal that updates the policy via GRPO.}
\label{fig:framework} 
\vspace{-12pt}
\end{figure*}

\subsection{Framework Overview}
We introduce a two-stage training framework for LVLMs that integrates the auxiliary-line construction into the reasoning process.
In the first stage, we apply SFT on automatically synthesized chain-of-thought (CoT) data with explicit auxiliary-line steps, enabling the model to actively construct auxiliary lines, thereby establishing a good initialization.
In the second stage, we further use RL to encourage the model to construct auxiliary lines that faithfully reflect the geometry of the diagram, boosting the precision of the solution.
At the core is a \textbf{cross-modal reward model} that provides fine-grained feedback by scoring the agreement between the original diagram plus the generated auxiliary-line description and a reference diagram annotated with the correct auxiliary line.
In summary, our framework combines direct supervision with structured visual feedback, thereby enhancing the reliability of auxiliary-line constructions and overall geometric problem-solving performance.

\subsection{Supervised Fine-Tuning }
We apply SFT on CoT exemplars to initialize the model for RL. Auxiliary-line construction steps are explicitly marked with \texttt{[AUX]} and \texttt{[/AUX]} to provide structured supervision and support later reward modeling, and the model is trained using a standard next-token prediction objective.



\subsection{Cross-Modal Reward-Driven RL}


As shown in Figure~\ref{fig:pilot_study}, accurate auxiliary-line constructions improve reasoning success. 
The key challenge is how to integrate such constructions into the reasoning process. Recent methods adopt Python-based drawing code to render auxiliary lines for intermediate reasoning, yet their reliance on precise and executable specifications limits robustness in complex solid geometry. Motivated by this limitation, and inspired by recent advances in textual CoT supervision~\citep{COS, CoT_based_Synthesizer}, we represent auxiliary-line constructions as structured natural language descriptions.
To bridge the gap between textual descriptions and spatial structure, we propose a \textbf{cross-modal reward model} that scores diagram-text alignment between the original diagram paired with a generated auxiliary-line description and a reference diagram annotated with the ground-truth auxiliary lines.
This reward provides geometry-aware supervision without relying on executable drawing code or precise geometric specifications, enabling robust scaling across diverse diagram styles.
We integrate this cross-modal reward, alongside a final-answer reward, into a GRPO-based RL stage to align intermediate constructions with the diagram while maintaining final-answer accuracy.
An overview of the stage is illustrated in Figure~\ref{fig:framework}.

\subsubsection{Cross-modal Reward Model}
\label{sec: cross_reward}

Given an original diagram $I$, a textual description of auxiliary lines $d$ (either the ground-truth description $d_{\text{aux}}$ or a perturbed variant $d^{-}_{\text{aux}}$), and a reference diagram $I^{+}$ annotated with the ground-truth auxiliary lines, the reward model evaluates the geometric relations implied by applying $d$ to $I$ against the additional geometric structures present in $I^{+}$ but absent from $I$.
Rather than relying on surface-level similarity, it assesses diagram-text relational consistency by verifying whether geometric relations specified in $d$, such as parallelism, orthogonality, and angle bisection, are satisfied in the reference diagram $I^{+}$.
Accordingly, the proposed reward exhibits three desirable properties:
\textbf{(1) Cross-modal relational alignment}, which evaluates diagram-text consistency at the level of geometric relations rather than surface-level or lexical similarity;
\textbf{(2) Render-free and coordinate-free}, as it avoids dependence on executable drawing code or pixel-accurate coordinates, relying instead on relational correspondence between diagrams and textual descriptions; and
\textbf{(3) Fine-grained supervision}, assigning intermediate scores to partially correct yet geometrically meaningful constructions, thereby enabling precise credit assignment in multi-step reasoning.
In summary, the cross-modal reward evaluates diagram-text spatial consistency at the level of geometric relations, assessing whether and to what extent the generated auxiliary lines align with the intended geometric structure, without relying on executable rendering code methods.
Next, we describe how to automatically \textbf{construct diagram-text supervision} at scale and use it to \textbf{train the cross-modal reward model}.



\vpara{Constructing Diagram-Text Supervision.} 
Each training example is represented as $\langle I, d, I^+, \hat{y} \rangle$, where $\hat{y}=(r,s)$ contains a brief justification $r$ and a consistency score $s \in [0, 1]$ indicating how well $d$ aligns with $I^{+}$ given $I$. 
We construct this supervision dataset via a fully automated pipeline (see Figure~\ref{fig:framework}).
Starting from high-quality supervision triplets $\langle I, d_{\text{aux}}, I^{+} \rangle$ constructed as described in Section~\ref{sec:data_creation}, we systematically generate challenging negative descriptions to support robust diagram-text supervision. Specifically, we design a set of rule-based perturbation templates that simulate common auxiliary-line construction errors, including \textit{partial deletion}, \textit{intersection alteration}, \textit{incorrect line connections}, \textit{adding irrelevant lines}, \textit{unrelated auxiliary lines}.
Building on these templates, we further leverage a large language model to synthesize diverse and linguistically varied negative descriptions $d^{-}_{\text{aux}}$. These negatives are lexically fluent and semantically plausible, yet geometrically inconsistent with the intended construction encoded in the reference diagram $I^{+}$, thereby providing hard counterexamples beyond surface-level mismatches.
To assess the consistency between each description \( d \in \{d_{\text{aux}}, d^{-}_{\text{aux}}\} \) and the target construction $I^{+}$, we adopt an \textit{LVLM-as-a-Judge} strategy. Concretely, the LVLM is prompted to evaluate diagram-text alignment by comparing the original diagram paired with the description $\langle I, d \rangle$ against the reference diagram $I^{+}$, and to produce both a natural-language rationale $r$ and a scalar alignment score $s \in [0,1]$.
This automated evaluation yields interpretable explanations alongside valued supervision signals, enabling scalable and fine-grained diagram-text supervision that spans faithful auxiliary-line descriptions as well as adversarially perturbed constructions.

\vpara{Training Cross-Modal Reward Model.} 
Given the input triplet $\langle I,d,I^{+}\rangle$, the model outputs a structured prediction $\hat{y}=(r,s)$.
We train the model by maximizing the conditional likelihood of the serialized output:
{
\vspace{-5pt}
\begin{equation}
\footnotesize
p_\phi(\hat{y} \mid I, d, I^+) = \prod_{i=1}^{T} p_\phi(\hat{y}_i \mid I, d, I^+, \hat{y}_{<i})
\end{equation}
\vspace{-12pt}
}

\noindent where $T$ denotes the length of the generated sequence $\hat{y}$, $\hat{y}_i$ is the $i$-th token in the output, and $\hat{y}_{<i}$ represents the sequence of previously generated tokens.
The consistency score is indicated as $r_{\text{aux}} = s = \text{Score}(\hat{y})$, where higher values correspond to better consistency.


Through this training, we obtain a reward model that provides precise feedback on diagram-text alignment between auxiliary-line descriptions and the reference diagram. 
This model serves as a key component of our framework, guiding the policy toward auxiliary-line constructions that are geometrically consistent and diagram-grounded.

\begin{figure*}[!t]
\centering 
\includegraphics[width=0.85\textwidth]{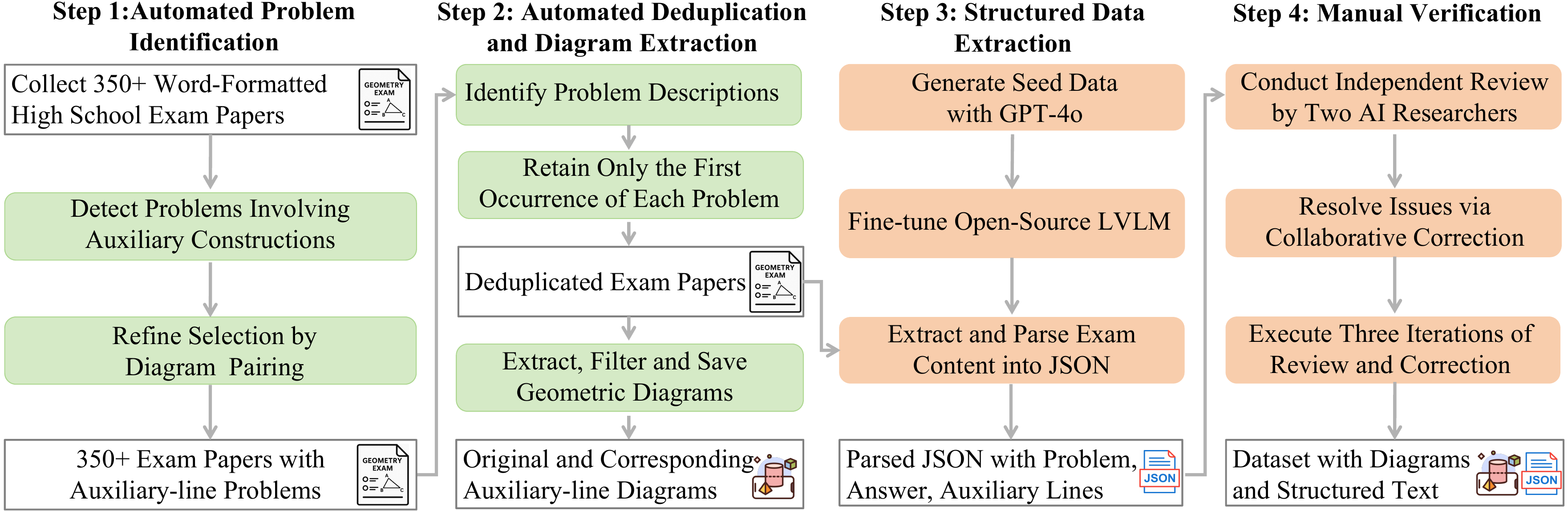}
\caption{Overview of the proposed data creation pipeline.}
\vspace{-15pt}
\label{fig:pipeline} 
\end{figure*}

\subsubsection{Optimization}

We adopt GRPO as the policy optimization algorithm. 
The overall reward signal combines the cross-modal reward with a final-answer reward defined as a binary score, yielding 1 if the predicted final answer matches the ground truth and 0 otherwise, \emph{i.e.}, $r = \alpha r_{\text{aux}} + (1-\alpha) r_{\text{ans}}$.

Given a geometric diagram $I$ and a question $q$, GRPO samples a set of response sequences $\{y_1, y_2, \dots, y_G\}$ from the old policy $\pi_{\theta_{\text{old}}}$. The policy $\pi_{\theta}$ is then optimized by maximizing the following objective:
{
\begin{equation}
\scriptsize
\begin{aligned}
\mathcal{L}_{\text{GRPO}} =
& \frac{1}{G} \sum_{i=1}^{G} \bigg(
    \min\bigg(
        \frac{\pi_\theta(y_i \mid I, q)}{\pi_{\theta_{\text{old}}}(y_i \mid I, q)} A_i,\ 
        \text{clip}\bigg(
            \frac{\pi_\theta(y_i \mid I, q)}{\pi_{\theta_{\text{old}}}(y_i \mid I, q)},\,
\\
& \qquad 1 - \epsilon,\ 1 + \epsilon
        \bigg) A_i
    \bigg)
    - \beta\, \mathbb{D}_{\text{KL}}(\pi_\theta \parallel \pi_{\text{ref}})
\bigg)
\end{aligned}
\label{eq:grpo}
\vspace{-5pt}
\end{equation}
}

Here, $G$ denotes the group size, while $\epsilon$ and $\beta$ are hyperparameters for clipping and the KL penalty. 

\section{Data Creation}
\label{sec:data_creation}
We curate the AuxSolidMath dataset to support training and evaluation of auxiliary-line reasoning in solid geometry.
As illustrated in Figure~\ref{fig:pipeline}, our data creation pipeline proceeds through four progressive steps: \textbf{automated problem identification, automated deduplication and diagram extraction, structured data extraction, and manual verification}. 
The pipeline standardizes raw exam problems into semantically aligned, high-quality multimodal instances that support training vision-language models for auxiliary-line geometric reasoning (More details in Appendix~\ref{app:data_creation}). 
Figure~\ref{fig:data_example} shows an example from the dataset. Each instance is represented as a five-tuple consisting of the \textit{problem description}, \textit{the final answer}, \textit{the auxiliary-line description}, \textit{the original diagram}, and \textit{the auxiliary-line diagram}.

\vpara{Automated Problem Identification.}
We curate AuxSolidMath from over 350 authentic high-school geometry exam sets using a two-stage filter that selects problems explicitly requiring auxiliary-line constructions, based on cue-verb retrieval from reference solutions and verification of paired original and auxiliary-line–annotated diagrams.

\vpara{Automated Deduplication and Diagram Extraction.}
We automatically deduplicate the dataset and extract paired diagrams to ensure unique and high-quality instances.
Duplicates are detected by textual matching, and only the first occurrence is retained.
For each retained problem, we extract the original and auxiliary-line-annotated diagrams and apply OpenCV-based filtering to discard low-resolution or unclear images.

\vpara{Structured Data Extraction.}
Building on the high-quality diagram pairs obtained in the previous step, we extract three textual fields for each problem: the problem description, the final answer, and the auxiliary-line description. Because the source Word files embed MathType formulas that standard parsers handle poorly, we render pages as images and parse them with a fine-tuned LVLM. All outputs are packaged in a structured JSON format.

\vpara{Manual Verification.}
Each instance is independently reviewed by two AI researchers to ensure accuracy, completeness, uniqueness, semantic consistency, and visual quality. 
Instances flagged by either reviewer undergo collaborative re-examination, with up to three review rounds to identify and correct subtle or ambiguous errors. 
This step resolves complex symbolic expressions and diagrammatic ambiguities that automated tools often misinterpret, and remains indispensable for ensuring high data fidelity in model training and evaluation.


\begin{figure*}[!t]
\centering 
\includegraphics[width=0.85\textwidth]{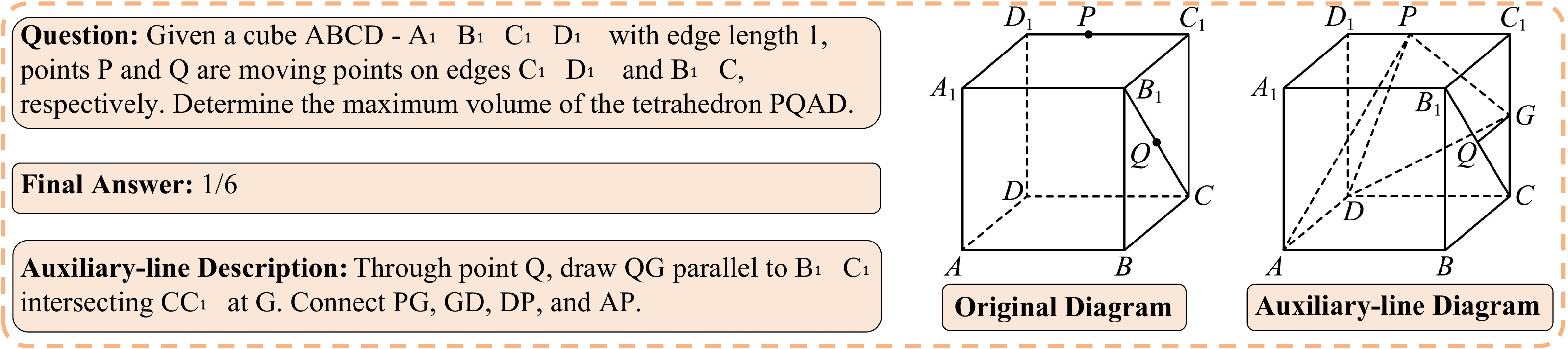}
\caption{An example from the AuxSolidMath dataset.}
\vspace{-15pt}
\label{fig:data_example} 
\end{figure*}

\vpara{Dataset Statistics.}
AuxSolidMath comprises 3,018 solid geometry problems collected from real high school examination papers. Within this dataset, we curate a new benchmark, GeoAuxBench, designed specifically to evaluate a model’s ability to construct auxiliary lines, a skill essential to solving complex geometry problems. GeoAuxBench contains 302 examples and is divided into two difficulty levels, \textit{Easy} (150) and \textit{Hard} (152), using the original difficulty annotations from the source exam papers rather than post hoc labels.
\textit{Hard} problems involve reasoning over implicit spatial relations (e.g., cross-plane or hidden projections), while \textit{Easy} problems rely on explicit relations. By inheriting these exam-defined criteria, GeoAuxBench supports realistic evaluation of auxiliary-line construction in solid geometry.

\section{Experiments}
\label{sec:exp}
\subsection{Experimental Settings}
\vpara{Benchmark.} We conduct our evaluation on \textbf{GeoAuxBench}, a newly introduced benchmark for auxiliary-line construction with two difficulty tiers: \textbf{Easy} and \textbf{Hard}.
We further report results on widely used benchmarks, \emph{i.e.}, \textbf{MathVision}~\citep{MATH-Vision}, \textbf{SolidGeo}~\citep{SolidGeo}, and the \textbf{OlympiadBench} subset of SolidGeo.

\vpara{Metrics.} We evaluate model performance using Pass@k~\citep{pass@k}, a widely adopted metric introduced by OpenAI. Specifically, we report Pass@1 and Pass@5: Pass@1 measures the accuracy of a single generated solution, while Pass@5 denotes the proportion of problems for which at least one of five generated solutions is correct.

\vpara{Models.} We assess \footnote{The code will be publicly available soon at https://github.com/PersistenceForever/GeoVLMath} at two model scales, 3B and 7B, both built upon the Qwen2.5-VL backbone.
We compare GeoVLMath against 20 strong LVLM baselines, covering both closed-source and open-source models, as well as representative methods following the code-driven visual construction paradigm, including V-Thinker~\cite{v-thinker} and CodePlot-CoT~\cite{CodePlot-CoT}.

\begin{table*}[t]
\centering
\setlength{\tabcolsep}{3pt}
\scalebox{0.8}{
\begin{threeparttable}
\vspace{0.5em} 
\newcolumntype{?}{!{\vrule width 1 pt}}
\renewcommand\arraystretch{0.6}
\begin{tabular}{@{}c@{ }?@{ }cc@{ }?@{ }cc@{ }?@{ }cc@{}}
\toprule
\multirow{2}{*}{\textbf{LVLM}} &
\multicolumn{2}{@{}c@{ }?@{ }}{\textbf{Easy}} &
\multicolumn{2}{@{}c@{ }?@{ }}{\textbf{Hard}} &
\multicolumn{2}{@{}c}{\textbf{Average}}\\
& \textbf{Pass@1}& \textbf{Pass@5} & \textbf{Pass@1}& \textbf{Pass@5}& \textbf{Pass@1}& \textbf{Pass@5}\\
\midrule
\multicolumn{7}{c}{\textbf{Closed-source LVLMs}}\\
\midrule
gpt-5-mini & 89.33 & 92.67 & 63.16 &75.00  &76.16 &83.78 \\
o4-mini-2025-04-16 & 84.00 & 93.33 & 60.53 & 74.34 & 72.19& 83.77\\
GPT-4o & 8.67& 25.33 & 6.58 & 15.13 &7.62 & 20.20\\
Gemini-2.0-Flash & 37.33 & 62.67 & 25.00 & 39.47 & 31.12 & 50.99\\
Gemini-2.5-Flash & 84.00 & 91.33 & 63.16 & 78.95 & 73.51 & 85.10 \\
Claude 3.7 Sonnet 20250219  & 15.33 & 41.33 & 13.16 & 28.29 &14.24 &34.77\\
Claude Sonnet 4 20250514 & 56.00 & 77.33 & 30.92 & 44.74 & 43.38 & 60.93\\
\midrule
\multicolumn{7}{c}{\textbf{Open-source LVLMs (3B-14B)}}\\
\midrule
V-Thinker (7B) & 8.00& 17.33 & 3.29& 10.53 & 5.63& 13.91\\
InternVL3-8B & 9.33 & 25.33 & 5.92 & 15.79 & 7.61 & 20.53 \\
Llama-3.2-11B-Vision-Instruct & 2.00 & 12.00 & 3.29 & 5.92 & 2.65 & 8.94\\
InternVL3-14B & 13.33 & 28.67 & 5.92 & 15.13 & 9.60 & 21.86\\
\midrule
Qwen2.5-VL-3B-Instruct & 2.00 & 14.89 & 1.97 & 8.33 & 1.98 & 11.59 \\
\textbf{GeoVLMath-3B (Ours)} & 12.89 & 20.44 & 5.70 & 9.87 & 9.27 & 15.12 \\
\midrule
Qwen2.5-VL-7B-Instruct & 5.14 & 20.67 & 3.95 & 11.18 & 4.54 & 15.89 \\
\textbf{GeoVLMath-7B (Ours)} & \textbf{14.67} & \textbf{35.56} & \textbf{5.92}& \textbf{16.67} & \textbf{10.27} &\textbf{26.05}\\
\midrule
\multicolumn{7}{c}{\textbf{Open-source LVLMs (17B-78B)}}\\
\midrule
CodePlot-CoT (32B) &4.00 &8.67 &2.63 &5.26 &3.31 &6.95\\
Qwen2-VL-72B-Instruct & 6.00 & 15.33 & 5.26 & 8.55& 5.63 & 11.92\\
Qwen2.5-VL-32B-Instruct & 20.67 & 23.33 & 11.18 & 13.16 & 15.89 & 18.21 \\
Llama-4-Scout-17B-16E-Instruct & 20.67 & 36.67 & 7.89 & 18.42 & 14.24 & 27.48\\
InternVL3-38B & 19.33 & 41.33 & 10.53 & 21.71 & 14.90 & 31.46\\
Qwen2.5-VL-72B-Instruct & 24.00 & 40.67 & 13.16 & 19.74 & 18.54 & 30.14 \\
InternVL3-78B & 16.67 & 36.67 & 9.21 & 21.05 & 12.92 & 28.81 \\
\bottomrule
\end{tabular}
\begin{tablenotes}
\item[*] Bold indicates the best results for models of similar sizes.
\end{tablenotes}
\vspace{-5pt}
\end{threeparttable}
}
\caption{Overall evaluation on GeoAuxBench (\%). }
\label{tb:overall_evaluation}
\vspace{-12pt}
\end{table*}

\begin{table*}[t]
\centering
\setlength{\tabcolsep}{3pt}
\scalebox{0.8}{
\begin{threeparttable}
\vspace{0.5em} 
\newcolumntype{?}{!{\vrule width 1.0pt}}
\renewcommand\arraystretch{0.6}
\begin{tabular}{@{}c@{ }?@{ }cc@{ }?@{ }cc@{ }?@{ }cc@{}}
\toprule
\multirow{2}{*}{\textbf{LVLM}} &
\multicolumn{2}{@{}c@{ }?@{ }}{\textbf{MathVision}} &
\multicolumn{2}{@{}c@{ }?@{ }}{\textbf{OlympiadBench}} &
\multicolumn{2}{@{}c}{\textbf{SolidGeo}} \\
& \textbf{Pass@1}& \textbf{Pass@5} & \textbf{Pass@1}& \textbf{Pass@5}& \textbf{Pass@1}& \textbf{Pass@5}\\
\midrule
V-Thinker (7B) &16.36 &  31.82& 4.60 &13.22 &10.39 &29.38\\
CodePlot-CoT (32B) & 10.00 &29.09 & 2.87 & 5.17 &16.62 & 30.86 \\
\midrule
Qwen2.5-VL-3B-Instruct &5.45  &  16.36&2.30  &6.32& 5.04 &12.17\\
\textbf{GeoVLMath-3B (Ours)} & 5.45 & 20.91 & 14.94& 25.86  &14.54 &32.34\\
\midrule
Qwen2.5-VL-7B-Instruct & 11.82 & 20.91 & 2.87 &13.22 & 5.34 & 14.84\\
\textbf{GeoVLMath-7B (Ours)} & \textbf{16.36} & \textbf{36.36} & \textbf{25.86}& \textbf{50.57}  & \textbf{17.51} &\textbf{42.34}\\
\bottomrule
\end{tabular}
\end{threeparttable}
}
\vspace{-6pt}
\caption{Overall evaluation on widely used benchmarks (\%). }
\label{tb:overall_evaluation2}
\vspace{-15pt}
\end{table*}

\subsection{Main Results}
As shown in Table~\ref{tb:overall_evaluation} and Table~\ref{tb:overall_evaluation2}, we make three key observations: \textbf{(1) GeoVLMath achieves strong and robust performance among models of comparable scale across benchmarks.}
On GeoAuxBench, both GeoVLMath-3B and GeoVLMath-7B outperform their corresponding base models, Qwen2.5-VL-3B-Instruct and Qwen2.5-VL-7B-Instruct, on pass@5. GeoVLMath-3B improves by \textbf{+3.53\%} (11.59\% $\rightarrow$ 15.12\%), while GeoVLMath-7B yields a larger gain of \textbf{+10.16\%} (15.89\% $\rightarrow$ 26.05\%). 
Moreover, GeoVLMath consistently surpasses the representative code-driven visual construction method V-Thinker on Pass@5, improving by \textbf{+12.14\%} (13.91\%  $\rightarrow$ 26.05\%), despite not relying on executable drawing code or intermediate diagram rendering.
Beyond this task-aligned benchmark, GeoVLMath also achieves the best performance among the compared models on widely used public benchmarks. These results demonstrate that the auxiliary-line-aware training signal generalizes beyond GeoAuxBench and improves geometric reasoning robustness.
\textbf{(2) GeoAuxBench-Hard is a challenging benchmark that clearly distinguishes LVLM capabilities in geometric reasoning.}
For example, even strong closed-source models such as Gemini-2.5-Flash and gpt-5-mini achieve Pass@1 scores of \textbf{63.16\%}, whereas Qwen2.5-VL-72B-Instruct attains only \textbf{13.16\%}, highlighting a substantial performance gap.
This is primarily because the benchmark demands deliberate auxiliary-line construction and multi-step spatial reasoning while minimizing shortcut opportunities, and provides reference diagrams for fine-grained error analysis.
Together, GeoAuxBench-Hard serves as a concise yet highly discriminative testbed for differentiating LVLM capabilities.
\textbf{(3) Model scale alone does not compensate for insufficient auxiliary-line awareness.} On GeoAuxBench-Easy, GeoVLMath-7B achieves higher pass@5 than Qwen2.5-VL-32B-Instruct (\textbf{35.56\%} vs. \textbf{23.33\%}).
On GeoAuxBench-Hard, GeoVLMath-7B also outperforms Qwen2.5-VL-32B-Instruct (\textbf{16.67\%} vs. \textbf{13.16\%}).
Error analysis shows that Qwen2.5-VL-32B-Instruct rarely constructs auxiliary lines and thus overlooks latent spatial constraints, whereas GeoVLMath-7B proactively introduces appropriate auxiliary lines and exploits the induced constraints during reasoning. This comparison highlights that auxiliary-line supervision, rather than model scale alone, is critical for reliable geometric reasoning.

\subsection{Cross-modal  Reward Model}
Leveraging AuxSolidMath triplets $\langle I, d_{\text{aux}}, I^{+}\rangle$, we apply rule-based perturbations to simulate typical auxiliary-line errors and use the resulting data to train a cross-modal reward model on Qwen2.5-VL-7B.
The dataset comprises 2,970 training examples and 330 test examples. We train the model for 3 epochs with a batch size of 16, using the AdamW optimizer with a learning rate of 2e-5 and a cosine learning rate scheduler with a 0.1 warm-up ratio. During training, the vision tower and projection modules are frozen, while the language model remains trainable. 
The model achieves a pass@1 accuracy of \textbf{98.18\%} on the test set, indicating reliable alignment between textual auxiliary-line descriptions and their visually annotated counterparts. 

\begin{table*}[t]
\centering
\setlength{\tabcolsep}{3pt}
\vspace{0.5em} 
\newcolumntype{?}{!{\vrule width 1.0pt}}
\renewcommand\arraystretch{0.6}
\scalebox{0.8}{
\begin{tabular}{@{}c@{ }?@{ }cc@{ }?@{ }cc@{ }?@{ }cc@{}}
\toprule
\multirow{2}{*}{} &
\multicolumn{2}{@{}c@{ }?@{ }}{\textbf{Easy}} &
\multicolumn{2}{@{}c@{ }?@{ }}{\textbf{Hard}} &
\multicolumn{2}{@{}c}{\textbf{Average}}\\
& \textbf{Pass@1}& \textbf{Pass@5} & \textbf{Pass@1}& \textbf{Pass@5}& \textbf{Pass@1}& \textbf{Pass@5}\\
\midrule
GeoVLMath-7B & 14.67 & 35.56 & 5.92& 16.67 & 10.27 &26.05\\
\midrule
w/o Cross-Modal Reward 
&10.89\textsubscript{$\scriptsize\downarrow$\textbf{3.78}}
& 32.22\textsubscript{$\scriptsize\downarrow$\textbf{3.34}}
& 4.82\textsubscript{$\scriptsize\downarrow$\textbf{1.10}}
& 13.60\textsubscript{$\scriptsize\downarrow$\textbf{3.07}}
& 7.83\textsubscript{$\scriptsize\downarrow$\textbf{2.44}}
& 22.85\textsubscript{$\scriptsize\downarrow$\textbf{3.20}}\\
Textual Reward 
&10.67\textsubscript{$\scriptsize\downarrow$\textbf{4.00}}
& 28.44\textsubscript{$\scriptsize\downarrow$\textbf{7.12}}
& 4.39\textsubscript{$\scriptsize\downarrow$\textbf{1.53}}
& 12.50\textsubscript{$\scriptsize\downarrow$\textbf{4.17}}
& 7.51\textsubscript{$\scriptsize\downarrow$\textbf{2.76}}
& 20.42\textsubscript{$\scriptsize\downarrow$\textbf{5.63}}\\
\midrule
w/o RL  
& 3.33\textsubscript{$\scriptsize\downarrow$\textbf{11.34}}
& 20.44\textsubscript{$\scriptsize\downarrow$\textbf{15.12}}
& 3.95\textsubscript{$\scriptsize\downarrow$\textbf{1.97}}
& 11.18\textsubscript{$\scriptsize\downarrow$\textbf{5.49}}
& 3.64\textsubscript{$\scriptsize\downarrow$\textbf{6.63}}
& 15.78\textsubscript{$\scriptsize\downarrow$\textbf{10.27}}\\
\bottomrule
\end{tabular}
}
\vspace{-6pt}
\caption{Results of ablation studies (\%). }
\label{tb:ablation_evaluation}
\vspace{-15pt}
\end{table*}

\subsection{Ablation Studies}
\vpara{Cross-Modal Reward.} We assess the role of cross-modal supervision with two variants, keeping all other settings unchanged.
(a) \textit{w/o cross-modal reward.}
This variant removes all supervision related to auxiliary lines and trains the model solely for final-answer accuracy, with no supervision on whether auxiliary lines are introduced. This allows us to assess the effect of training with answer-only supervision, approximating a setting where auxiliary lines are omitted from the training objective.
(b) \textit{Textual reward.}
In this variant, cross-modal consistency is replaced with a text-only semantic similarity objective that evaluates the similarity between the generated auxiliary-line description and the ground-truth annotation. Concretely, we use \emph{EmbeddingGemma}~\citep{embeddinggemma2025} to encode sentences and compute a similarity score for training. 
This variant favors fluent textual descriptions but does not enforce grounding to the input diagram.
\textit{Findings.}
As reported in Table~\ref{tb:ablation_evaluation}, removing the cross-modal reward results in performance degradation (Average pass@1: \textbf{10.27 $\rightarrow$ 7.83}, pass@5: \textbf{26.05 $\rightarrow$ 22.85}), underscoring the importance of geometry-aware supervision for instances that require introducing auxiliary lines.
Substituting it with a purely textual similarity objective performs even worse (Average pass@1: \textbf{10.27 $\rightarrow$ 7.51}, pass@5: \textbf{26.05 $\rightarrow$ 20.42}), consistent with our pilot finding in Section~\ref{sec:intro} that incorrect auxiliary lines can be worse than none.
These declines suggest that lexical alignment introduces spurious signals and conflicts with precise diagram grounding, favoring surface-level paraphrases over geometry-aware reasoning.
Error analysis reveals distinct failure modes: (a) often ignores auxiliary-line construction and overfits to answer-only cues; (b) produces fluent but visually inconsistent descriptions (e.g., incorrect lines) that fail to constrain diagram-based reasoning. 
Overall, text-only alignment fails to faithfully capture geometric structure, motivating the need for visually grounded, structure-preserving diagram-text alignment to support rigorous verification of geometric measurements and structural relations.


\vpara{Reinforcement Learning.} 
To quantify the contribution of reinforcement learning, we remove the RL stage and train an SFT-only variant. As shown in Table~\ref{tb:ablation_evaluation}, GeoVLMath-7B trained with SFT+RL consistently outperforms its SFT-only counterpart, with clear performance degradation when the RL stage is removed.
This improvement reflects the role of RL in moving beyond strict imitation: reward-aligned optimization encourages exploration of more effective strategies and enables credit assignment for beneficial intermediate steps, rather than relying on surface-level matching alone. As a result, RL acts as a post-SFT catalyst that consolidates preliminary SFT competence into robust multi-step reasoning, particularly in scenarios requiring auxiliary-line construction.
\section{Related Work}
\label{sec:related_work}

Recent LVLMs~\citep{claude4,Gemini2.5flash,o3} have advanced geometric problem solving, particularly in plane geometry. Prior work mainly follows two lines: direct generation of answers or reasoning paths from multimodal inputs~\citep{GNS,GeoX,G-LLaVA}, which is constrained by intrinsic reasoning capacity, and tool-augmented reasoning that generates executable code for symbolic computation or geometric operations~\citep{Pi_GPS,GeoCoder,DCM}.
More recently, a code-driven visual construction paradigm has emerged, where models generate code to construct or modify diagrams during reasoning, as exemplified by Visual Sketchpad~\citep{Visual_Sketchpad}, V-Thinker~\citep{v-thinker}, and CodePlot-CoT~\citep{CodePlot-CoT}. 
While these methods enable interactive visual feedback, they depend on precise code execution and explicit coordinate annotations, and tightly couple auxiliary-line construction with the LVLM's intrinsic reasoning, thereby limiting robustness in solid geometry.
In contrast, our approach incorporates auxiliary-line construction into an RL framework guided by a cross-modal reward model, decoupling construction quality from intrinsic reasoning, without explicit geometric specifications, enabling robust auxiliary-line reasoning across model scales.
\section{Conclusion}
\label{sec:con}

Auxiliary-line reasoning in solid geometry exposes a key limitation of current LVLMs: the lack of reliable learning signals for structured visual construction. We address this challenge by formulating auxiliary-line construction as a vision-language alignment problem and introducing a RL framework guided by a cross-modal reward that directly measures diagram-text correspondence. This design enables stable optimization without relying on explicit geometric specifications or executable code. To support learning at scale, we automatically construct \textbf{AuxSolidMath}, a high-quality dataset of real-exam solid geometry problems with aligned diagrams and auxiliary-line annotations, providing a reusable resource for the community. Leveraging this framework, \textbf{GeoVLMath} consistently improves auxiliary-line reasoning.

\section*{Limitations}
\label{app:limit}
While our approach demonstrates strong effectiveness for auxiliary-line reasoning in solid geometry, our experiments primarily consider settings where auxiliary-line constructions admit concise high-level descriptions. Extending the evaluation to even broader or more diverse geometric constructions remains an interesting direction for future work.
In addition, our cross-modal reward model relies on annotated auxiliary-line diagrams for supervision during reinforcement learning. Although we introduce AuxSolidMath to support this setting, exploring more scalable forms of weak or self-supervised reward signals is a promising direction for future research.
\bibliography{custom}

\appendix
\section{The Use of AI Assistants}
In this paper, ChatGPT was used exclusively for language polishing, including grammar correction, phrasing, and stylistic refinement. It was not used to generate scientific content such as research ideas, methods, experiments, or related work. No confidential, personal, or proprietary information was shared with the model. The authors take full responsibility for the scientific content, which was entirely authored and verified by the authors.

\section{Data Creation}
\label{app:data_creation}
In this section, we detail the four progressive steps of our data creation pipeline.
\subsection{Automated Problem Identification}
To construct the AuxSolidMath dataset, we first collect over 350 sets of high school geometry problems from publicly available online sources. 
Given that the dataset is intended to support constructive geometric reasoning, we specifically target problems that necessitate auxiliary line constructions as integral components of their solutions.

To efficiently identify such problems, we design an automated two-stage filtering pipeline using Python scripts. In the first stage, we detect problems whose solutions contain explicit mentions of auxiliary-line constructions. Specifically, we apply regular expression patterns to locate question number markers that are explicitly present in the exam papers and use these markers to segment the content into individual problem units. 
For each problem, we examine the solution for verbs that signal the introduction of auxiliary lines (e.g., "connect," "construct," "draw," "establish").
Problems lacking such terms are discarded, while those containing relevant cues are retained. In the second stage, we further refine the selection by ensuring that each retained problem contains both the original diagram and an auxiliary-line diagram. To this end, we quantify the number of diagrams associated with each problem. Problems with fewer than two diagrams are excluded, whereas those with at least two, which usually represent the original and modified diagrams, are preserved. This automated pipeline enables scalable and consistent filtering of auxiliary-line geometry problems, significantly reducing manual annotation effort.

\subsection{Automated Deduplication and Diagram Extraction}
Upon identifying geometry problems requiring auxiliary lines, we employ an automated pipeline to deduplicate instances and extract the associated diagrams. This step guarantees the uniqueness and visual quality of data instances for downstream model training.

\textbf{Problem Deduplication.} To eliminate duplicate problems, we retain only the first occurrence of each unique problem based on its textual content. Concretely, we initialize a global problem set as an empty collection. We then sequentially process all Word-formatted exam papers, examining only the problem descriptions while ignoring the associated solutions and diagrams. 
For each problem, if its description is not already present in the global set, we add the problem; otherwise, we discard it as a duplicate.
This procedure ensures that identical problems, which often recur across different examinations, are retained only once.

\textbf{Diagram Extraction.} Following deduplication, we extract, filter, and store the geometric diagrams associated with each retained problem. A key challenge lies in reliably distinguishing true geometric figures from image-embedded mathematical expressions (\emph{e.g.}, MathType equations), as both appear in Word exam papers. Existing Python libraries are unable to make this distinction accurately, often misclassifying equations as diagrams and introducing significant noise into the extraction process. To overcome this limitation, we innovatively integrate the Apache POI library through a custom Java implementation, enabling fine-grained control over the parsing of Word documents. This setup enables reliable identification and extraction of genuine geometric diagrams while effectively filtering out formula-rendered images. 
To further ensure visual quality, the extracted diagrams are then processed using OpenCV to discard low-resolution or unclear diagrams.
The remaining diagrams are subsequently saved using a standardized naming convention that distinguishes between the original and the annotated versions of the auxiliary lines. 
To be more specific, for each problem indexed by i, we store two images: \{i\}.png,  which contains the original diagram, and \{i\}\_auxiliary.png, which includes the corresponding auxiliary-line diagram. This consistent format facilitates downstream alignment between textual and visual modalities within the multimodal processing pipeline.

\subsection{Structured Data Extraction}
Building on the high-quality geometric diagrams obtained in the previous step, we proceed to extract the corresponding textual content for each instance, including the problem description, the final answer, and the auxiliary-line description. This extraction process is non-trivial, as the original Word documents frequently embed mathematical expressions in MathType formats that are not reliably supported by standard document parsing tools.

To address this challenge, we render the processed Word documents as images, thereby enabling LVLMs to leverage their visual reasoning capabilities. Although this approach appears straightforward, open-source models such as Qwen2.5-VL-7B-Instruct~\citep{Qwen2.5vl} often struggle to accurately parse complex geometry problems involving symbolic notation and mathematical expressions.
In contrast, closed-source models like GPT-4o~\citep{Gpt4o} exhibit significantly stronger performance, but their reliance on commercial APIs introduces substantial costs and limits scalability in large-scale applications.
To balance accuracy with scalability, we adopt a hybrid strategy\footnote{DeepSeek-OCR\cite{deepseekocr} was released on October 20, 2025 and was not available during data construction. A Qwen2.5-VL-based extractor fine-tuned on 300 seed samples achieved approximately 98\% accuracy and was therefore used.}. Specifically, we first utilize an advanced closed-source model (\emph{i.e.}, GPT-4o) to generate a small, high-quality seed dataset comprising 300 manually verified instances. This curated dataset is then used to fine-tune an LVLM (\emph{i.e.}, Qwen2.5-VL-7B-Instruct), resulting in a lightweight, domain-adapted model capable of accurate and scalable text extraction.
The final output consists of the extracted problem description, the final answer, and the auxiliary-line description, all encapsulated in a structured JSON format. This unified representation facilitates consistent data handling and serves as a foundation for training a robust open-source text extraction model. By releasing this model, we aim to contribute a practical and reusable resource to the broader research community working on geometry-aware vision-language understanding.

\subsection{Manual Verification}

To ensure the quality and reliability of the final dataset, we perform a manual verification step that assesses each data instance in terms of accuracy, completeness, uniqueness, and semantic consistency, alongside visual quality criteria such as image clarity and resolution. Two AI researchers serve as independent checkers. Each instance is independently reviewed by both researchers. If either checker identifies a potential issue, the instance is collaboratively revised. This process is repeated up to three times per instance, ensuring that all errors, including subtle or ambiguous ones, are systematically identified and corrected. Manual verification plays a critical role in resolving complex symbolic expressions and ambiguous diagrammatic content that automated tools may misinterpret. Despite its relatively low cost and effort, this step remains indispensable for ensuring the high data fidelity necessary for a reliable model.

\section{Experimental Setup}
\label{app:training}
\subsection{Models}
On the closed-source models, we include leading models such as gpt-5-mini~\citep{gpt5}, o4-mini~\citep{o3} and GPT-4o~\citep{Gpt4o}, Gemini-2.0-Flash and Gemini-2.5-Flash~\citep{Gemini2.5flash}, Claude 3.7 Sonnet~\citep{Claude3.7} and Claude Sonnet 4 20250514~\citep{claude4}. These models represent the forefront of multimodal reasoning among closed-source models, although their internal architectures remain undisclosed.
On the open-source models, we consider several publicly available high-performance models, including the Qwen2 VL~\citep{Qwen2vl} and Qwen2.5 VL series~\citep{Qwen2.5vl}, InternVL 3 families~\citep{InternVL3}, LLaMA-3.2-11B-Vision-Instruct~\citep{llama3.2} and Llama-4-Scout-17B-16E-Instruct~\citep{llama4}. 
In addition, we include representative methods following the code-driven visual construction paradigm, namely V-Thinker~\citep{v-thinker} and CodePlot-CoT~\citep{CodePlot-CoT}, which generate executable code to construct or modify diagrams as part of the reasoning process. These approaches provide an important comparison point for evaluating different strategies of integrating visual construction into multimodal reasoning.
Together, these models encompass a range of design paradigms, parameter scales, and instruction tuning strategies, forming a comprehensive and robust foundation for evaluating auxiliary-line reasoning in multimodal settings.
Note that models such as Gemini-2.5 Pro~\citep{Gemini2.5pro} and OpenAI o3~\citep{o3} are excluded from our study due to limited accessibility and high inference costs.

\subsection{Training Implementation Details}
We adopt a two-stage training paradigm based on the Qwen2.5-VL series, including Qwen2.5-VL-3B and Qwen2.5-VL-7B, consisting of the SFT stage and the RL stage.

\textbf{SFT Stage.} The SFT phase is conducted using the \texttt{LLaMA-Factory} framework~\citep{llama_factory}. For Qwen2.5-VL-7B, we train the model for 5 epochs with a per-device batch size of 2 and a gradient accumulation step of 8 (effective batch size of 16). We use the AdamW optimizer with a learning rate of 2e-5 and apply a cosine learning rate scheduler with a warmup ratio of 0.1. The model is trained in \texttt{bf16} precision. Vision and projection modules are frozen during this stage, while the language model remains unfrozen. For Qwen2.5-VL-3B, we adopt the same training configuration as the 7B variant, except learning rate and training epochs. Specifically, Qwen2.5-VL-3B is trained for 5 epochs with a learning rate of 3e-5.


\textbf{RL Stage.} 
For the Qwen2.5-VL-7B model, both training and validation data are loaded from Parquet files containing question-diagram pairs. The maximum response length is set to 8,192 tokens, and both the rollout and validation batch sizes are set to 16. The actor is optimized using AdamW (learning rate 2e-6, weight decay 1e-2, no warmup). KL regularization is applied using the \texttt{low\_var\_kl} penalty with a coefficient of 1e-2. Training runs for 6 epochs using \texttt{bf16} precision, with gradient checkpointing and partial FSDP offloading enabled for memory efficiency.

\emph{Rewards.} The overall reward is the sum of a cross-modal auxiliary-line consistency reward and a final-answer accuracy reward, where the auxiliary-line component is weighted by $\alpha$ = 0.1.
For Qwen2.5-VL-3B, we adopt the same RL configuration as the 7B model, with adjustments to the batch size and the number of training epochs. Specifically, Qwen2.5-VL-3B is trained for 4 epochs with a batch size of 8.

All training was conducted on a server equipped with two NVIDIA A100 80GB and two NVIDIA A800 80GB GPUs. The SFT stage was performed on two A100 GPUs, whereas the reinforcement learning stage utilized all four GPUs. For response generation during training and evaluation, we enabled stochastic sampling with temperature set to 0.7 and top-p set to 0.95.

\section{Related Work}
\label{app:related_work}
In this section, we also review benchmarks and datasets for geometric reasoning. 
Most benchmarks and datasets for geometric problem solving focus on plane geometry, where diagrams and problems involve two-dimensional figures. Well-known resources in this area include Geometry3K~\citep{Geometry3k}, GeoQA~\citep{GeoQA}, UniGeo~\citep{UniGeo}, and GeomRel~\citep{GeomRel}, which primarily cover plane geometry problems.
A concurrent benchmark, GeoLaux~\citep{GeoLaux}, explores the use of auxiliary lines in plane geometry, but is limited to simple cases and lacks engagement with the spatial complexity of solid geometry.
Nonetheless, there remains a lack of dedicated resources for solid geometry, even though solving such problems often requires interpreting three-dimensional relationships and drawing auxiliary lines to uncover hidden spatial structures.
While SolidGeo~\citep{SolidGeo} is a recent benchmark that focuses exclusively on solid geometry, it does not explicitly require auxiliary lines for solving its problems, leaving this important aspect of spatial reasoning underexplored.
Similarly, other benchmarks such as 
MathVision\citep{MATH-Vision}, MathVista~\citep{MathVista}, and MathVerse~\citep{MATHVERSE} contain only a limited number of solid geometry problems, and these also do not require auxiliary lines to reach the solution. As a result, these resources fall short of evaluating a model's ability to solve complex solid geometry problems where auxiliary lines are essential for uncovering implicit spatial relationships.
To address this gap, we present AuxSolidMath, the first dedicated dataset for solid geometry problems that require auxiliary lines to solve.
It offers comprehensive multimodal supervision, including the original diagram, the problem statement, textual descriptions of the required auxiliary lines, the final answer, and a corresponding diagram annotated with those lines, enabling models to learn how the auxiliary lines facilitate solid geometry reasoning.

\begin{figure*}[t]
\begin{tcolorbox}[colframe=gray!50!black,
  colback=gray!10!white,
  boxrule=1pt,
  left=5pt,right=5pt,
  fontupper=\footnotesize,
  width=\textwidth,
  title=SYSTEM\_PROMPT\_FOR\_SUPERVISED\_FINE-TUNING,
  fonttitle=\bfseries\color{green}\normalsize,
  boxed title style={
    left=2pt,right=2pt,top=1pt,bottom=1pt
  }]

\texttt{SYSTEM\_PROMPT\_FOR\_SUPERVISED\_FINE-TUNING} = """\\
You are a mathematician skilled in solving geometry problems through step-by-step reasoning. Solve the given geometry problem based on a geometric diagram and a natural language question. Use `[AUX]...[/AUX]' to indicate auxiliary constructions, such as establishing coordinate systems or constructing auxiliary lines. Finally, provide your final answer within `Final Answer:...'.

"""
\end{tcolorbox}

\vspace{0.5em}

\begin{tcolorbox}[colframe=gray!50!black,
  colback=gray!10!white,
  boxrule=1pt,
  left=5pt,right=5pt,
  fontupper=\footnotesize,
  width=\textwidth,
  title=USER\_PROMPT\_FOR\_SUPERVISED\_FINE-TUNING,
  fonttitle=\bfseries\color{green}\normalsize,
  boxed title style={
    left=2pt,right=2pt,top=1pt,bottom=1pt
  }]

\texttt{USER\_PROMPT\_FOR\_SUPERVISED\_FINE-TUNING} = """\\
Image: <image>\\
Question: \{\texttt{question}\}

"""
\end{tcolorbox}
\caption{Prompts for supervised fine-tuning.}
\label{fig:prompt_sft}
\end{figure*}

\begin{figure*}[!ht]
\begin{tcolorbox}[colframe=gray!50!black,
  colback=gray!10!white,
  boxrule=1pt,
  left=5pt,right=5pt,
  fontupper=\footnotesize,
  width=\textwidth,
  title=SYSTEM\_PROMPT\_FOR\_CROSS-MODAL\_REWARD\_MODEL,
  fonttitle=\bfseries\color{green}\normalsize,
  boxed title style={
    left=2pt,right=2pt,top=1pt,bottom=1pt
  }]

\texttt{SYSTEM\_PROMPT\_FOR\_CROSS-MODAL\_REWARD\_MODEL} = """\\
You are a professional geometry reasoning evaluator. Your task is to evaluate whether a given textual description of auxiliary lines accurately explains the visual difference between the original diagram and the auxiliary-line diagram. \\

Score the description on a scale from 0 to 1:\\
1. 1 indicates a fully accurate and helpful description.\\
2. 0 indicates a completely irrelevant or misleading description.\\
3. Intermediate values (e.g., 0.25/ 0.50/ 0.75) reflect partial relevance or minor issues. \\

Return exactly one line: \\
<brief justification>. Score: <s>.

"""
\end{tcolorbox}

\vspace{0.5em}
\begin{tcolorbox}[colframe=gray!50!black,
  colback=gray!10!white,
  boxrule=1pt,
  left=5pt,right=5pt,
  fontupper=\footnotesize,
  width=\textwidth,
  title=USER\_PROMPT\_FOR\_CROSS-MODAL\_REWARD\_MODEL,
  fonttitle=\bfseries\color{green}\normalsize,
  boxed title style={
    left=2pt,right=2pt,top=1pt,bottom=1pt
  }]

\texttt{USER\_PROMPT\_FOR\_CROSS-MODAL\_REWARD\_MODEL} = """\\
Image (original diagram): <image $I$> \\
Image (auxiliary-line diagram): <image $I^+$>\\
Auxiliary-line description: \texttt{\{generated\_aux\_description\}} 

"""
\end{tcolorbox}
\caption{Prompts for cross-modal reward model.}
\label{fig:prompt_reward}
\end{figure*}

\section{Prompts}
\subsection{Prompts for Supervised Fine-tuning}
Figure~\ref{fig:prompt_sft} presents the two-part prompt template used in our supervised dataset. The system prompt assigns the solver role and enforces formatting: auxiliary lines must be wrapped in \texttt{[AUX]…[/AUX]} and the final answer must appear as plain text in \texttt{Final Answer:...}. The user prompt is multimodal, pairing a diagram referenced by the \texttt{<image>} token with the natural language question \texttt{\{question\}}, which yields explicit reasoning steps and a final answer.


\subsection{Prompts for Cross-Modal Reward Model}
Using the prompt as shown in Figure~\ref{fig:prompt_reward}, the cross-modal reward model compares the description of the auxiliary line generated by the policy model against a pair of diagrams, the original image $I$ and its auxiliary-line counterpart $I^+$, and returns a single line justification and a calibrated score in \([0,1]\) that measures visual–textual agreement. The instruction emphasizes the correctness of auxiliary-line constructions and adherence to geometric constraints. Higher scores indicate stronger alignment.

\section{Representative Examples}
\begin{figure*}[!t]
  \centering
  \includegraphics[width=\textwidth]{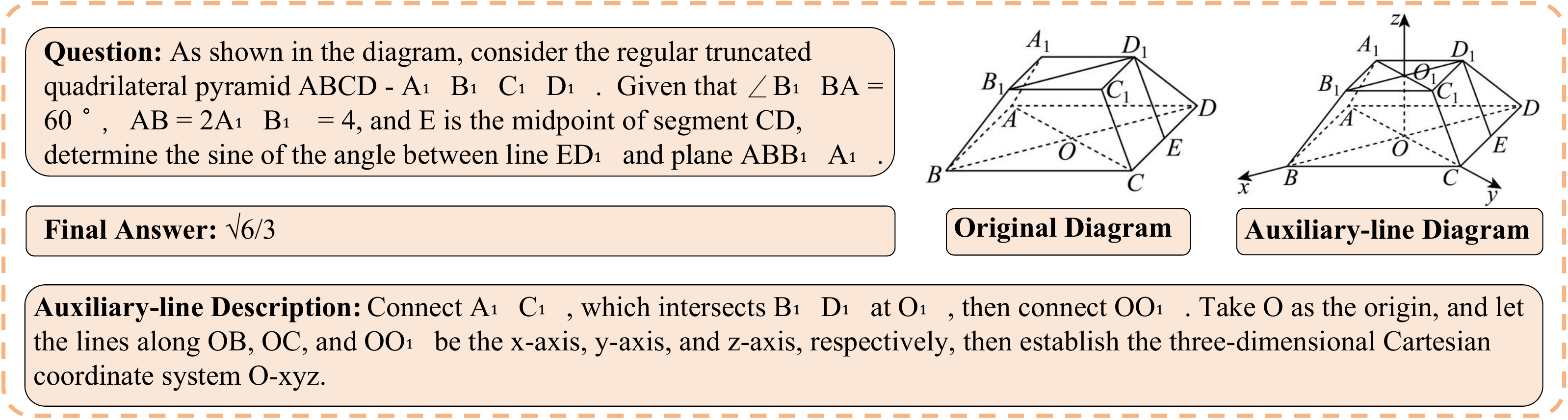}\par\vspace{1em}
  \includegraphics[width=\textwidth]{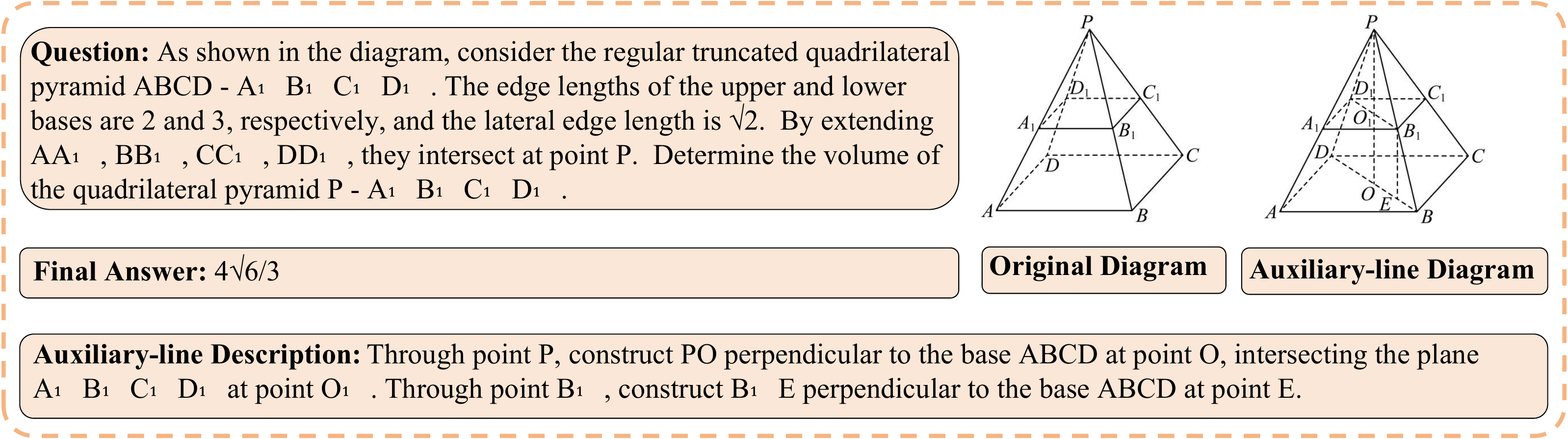}\par\vspace{1em}
  \includegraphics[width=\textwidth]{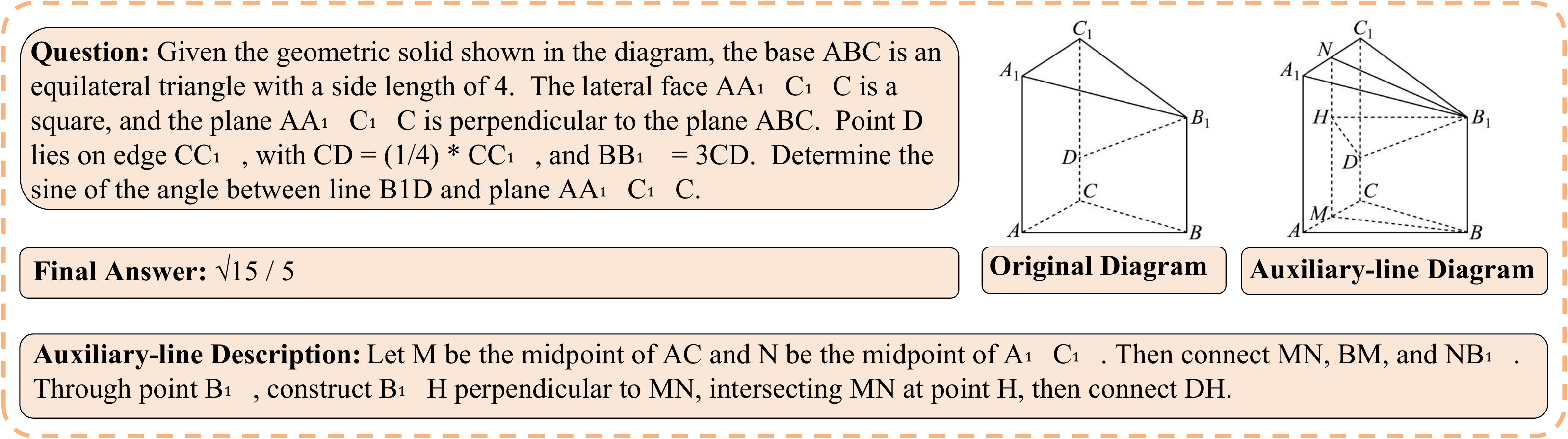}\par\vspace{1em}
  \includegraphics[width=\textwidth]{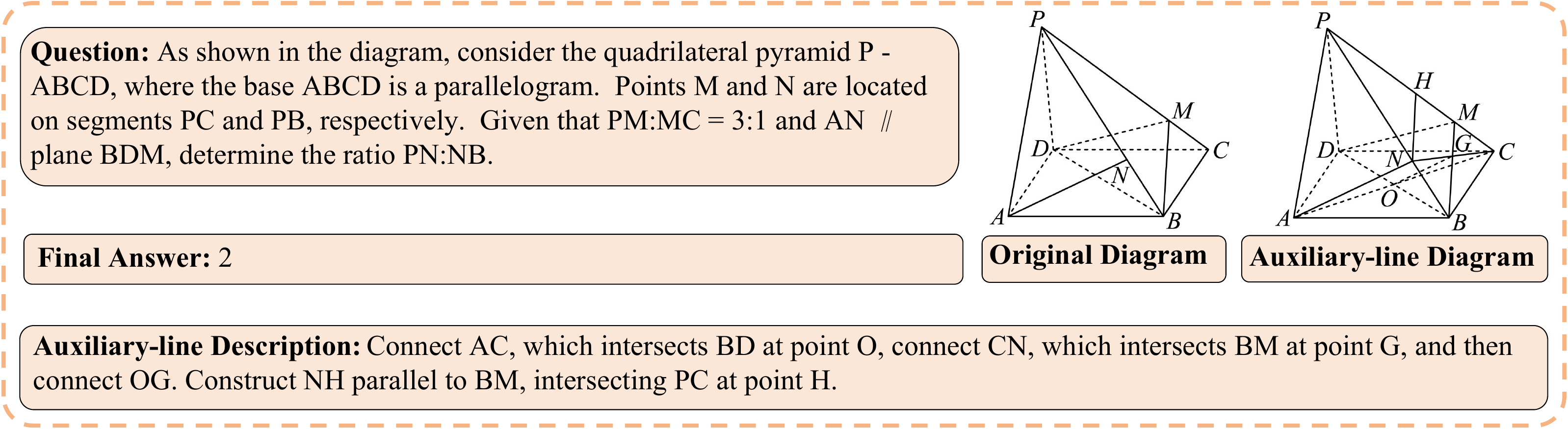}\par\vspace{1em}
  \includegraphics[width=\textwidth]{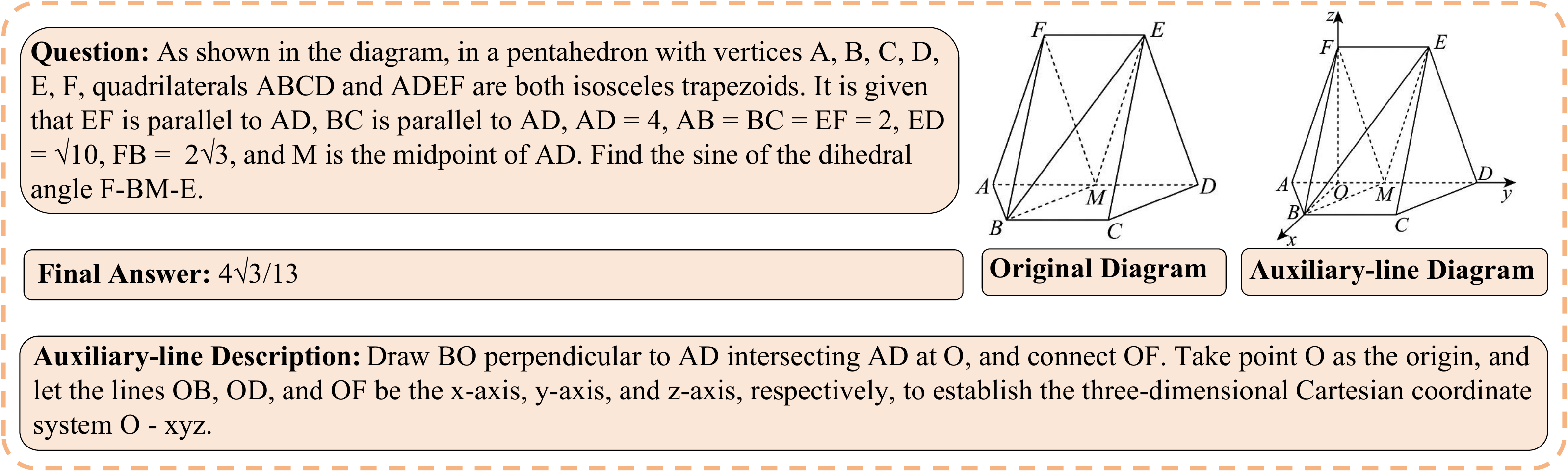}
  \caption{Representative examples from the AuxSolidMath dataset.}
  \label{fig:data_examples}
\end{figure*}

As illustrad in Figure~\ref{fig:data_examples} present qualitative examples from the AuxSolidMath dataset, including the question, the final answer, the auxiliary-line description, the original diagram, and the auxiliary-line diagram.
The examples showcase diverse strategies for constructing auxiliary lines and demonstrate that explicit annotations reveal the key spatial constraints.

\section{Case Study}
\subsection{Successful Cases}

\begin{figure*}[!ht]
  \centering
   \includegraphics[width=\textwidth]{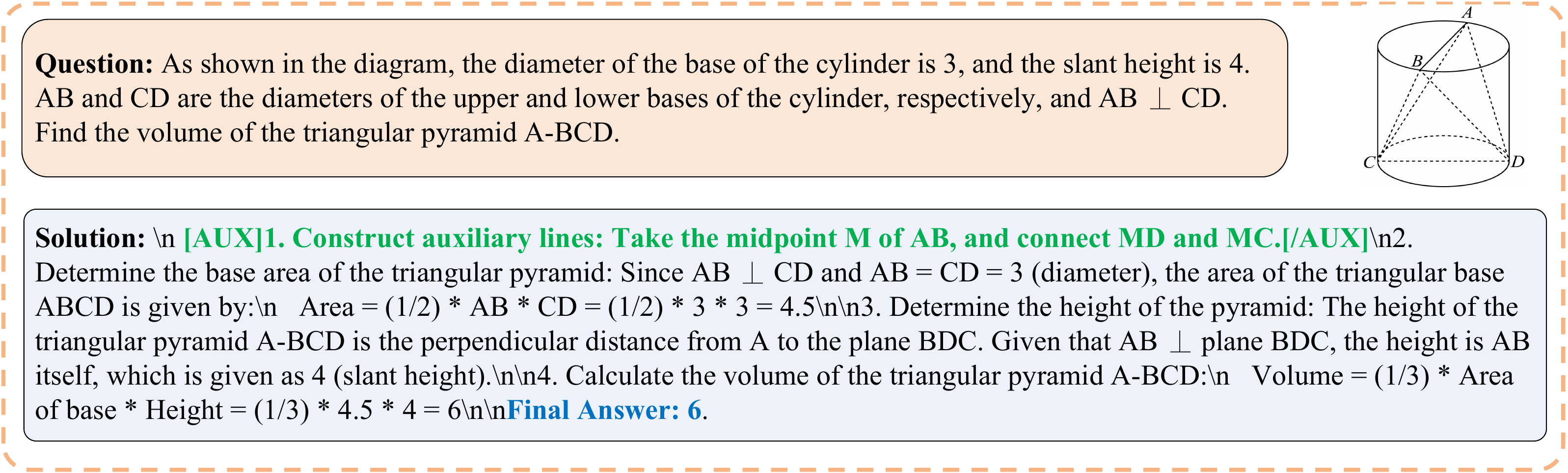}\par\vspace{1em}
   \includegraphics[width=\textwidth]{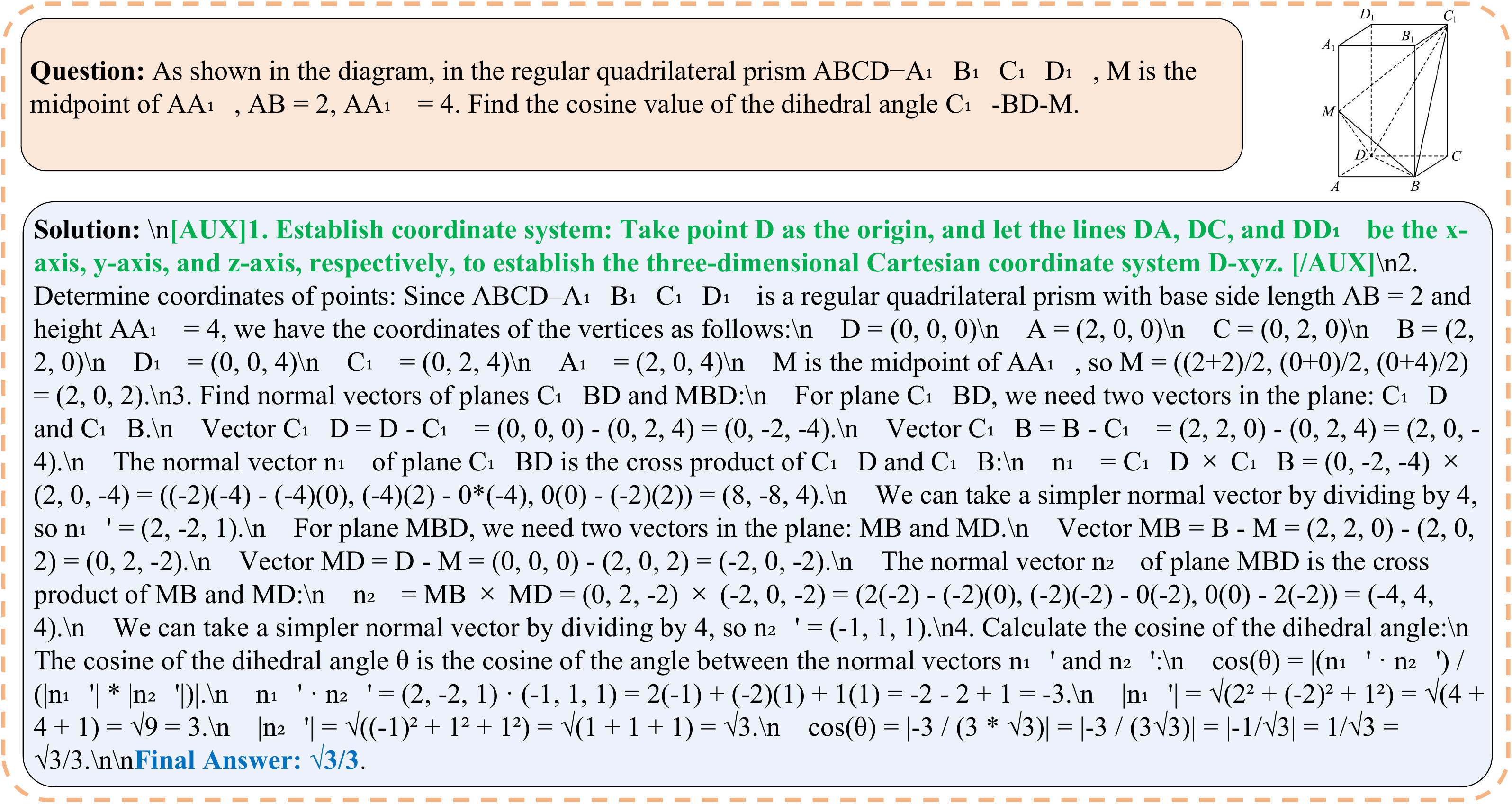}\par\vspace{1em}
  \includegraphics[width=\textwidth]{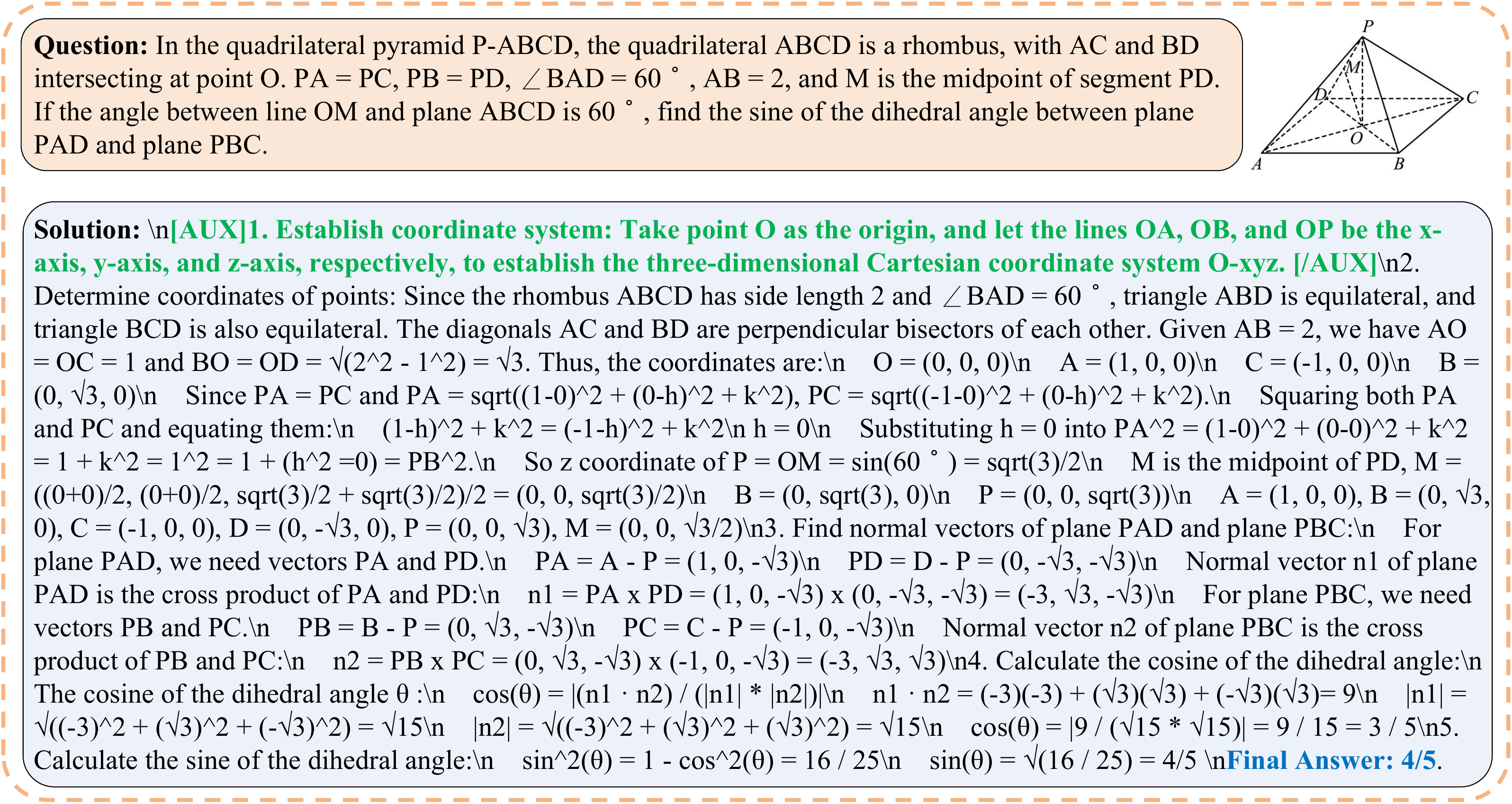}
  \caption{Successful cases generated by GeoVLMath-7B.}
  \label{fig:correct_cases}
\end{figure*}

In this section, we present representative cases produced by GeoVLMath-7B (see Figure~\ref{fig:correct_cases}). The examples illustrate how valid auxiliary-line constructions capture essential spatial constraints, which in turn lead to correct final answers. Collectively, these results provide concrete evidence of both the effectiveness of our training framework and the model’s capacity to tackle complex geometric reasoning requiring robust visual understanding and precise symbolic inference.

\begin{figure*}[h]
  \centering
   \includegraphics[width=\textwidth]{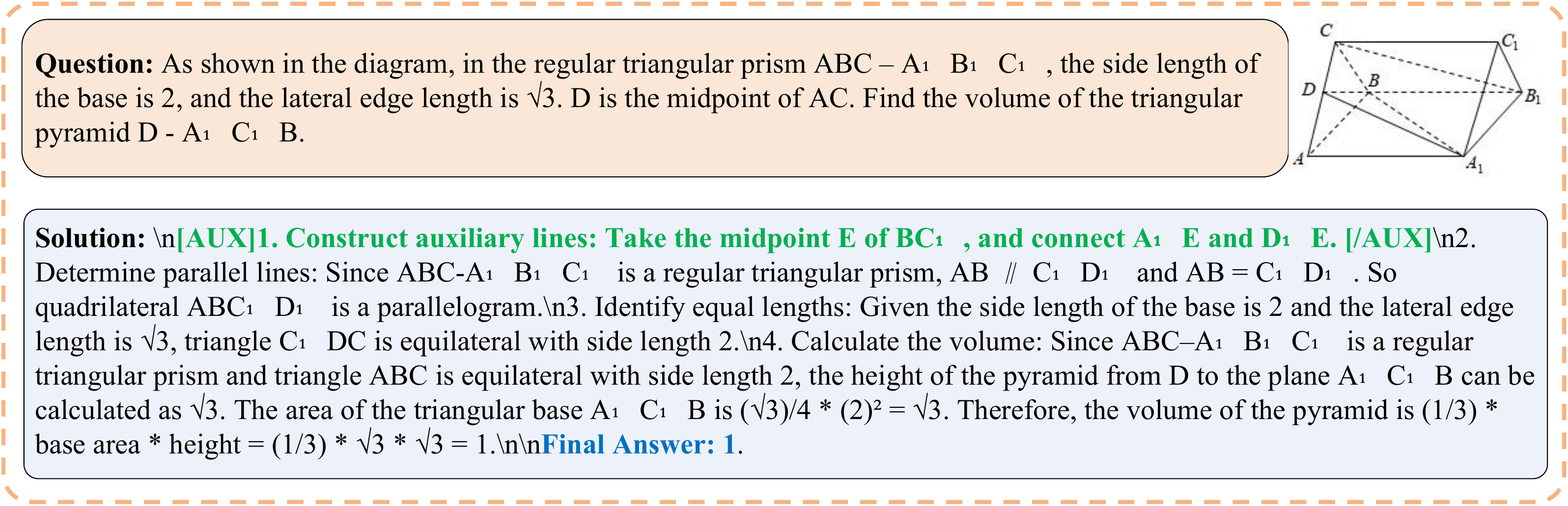}\par\vspace{2em}
      \includegraphics[width=\textwidth]{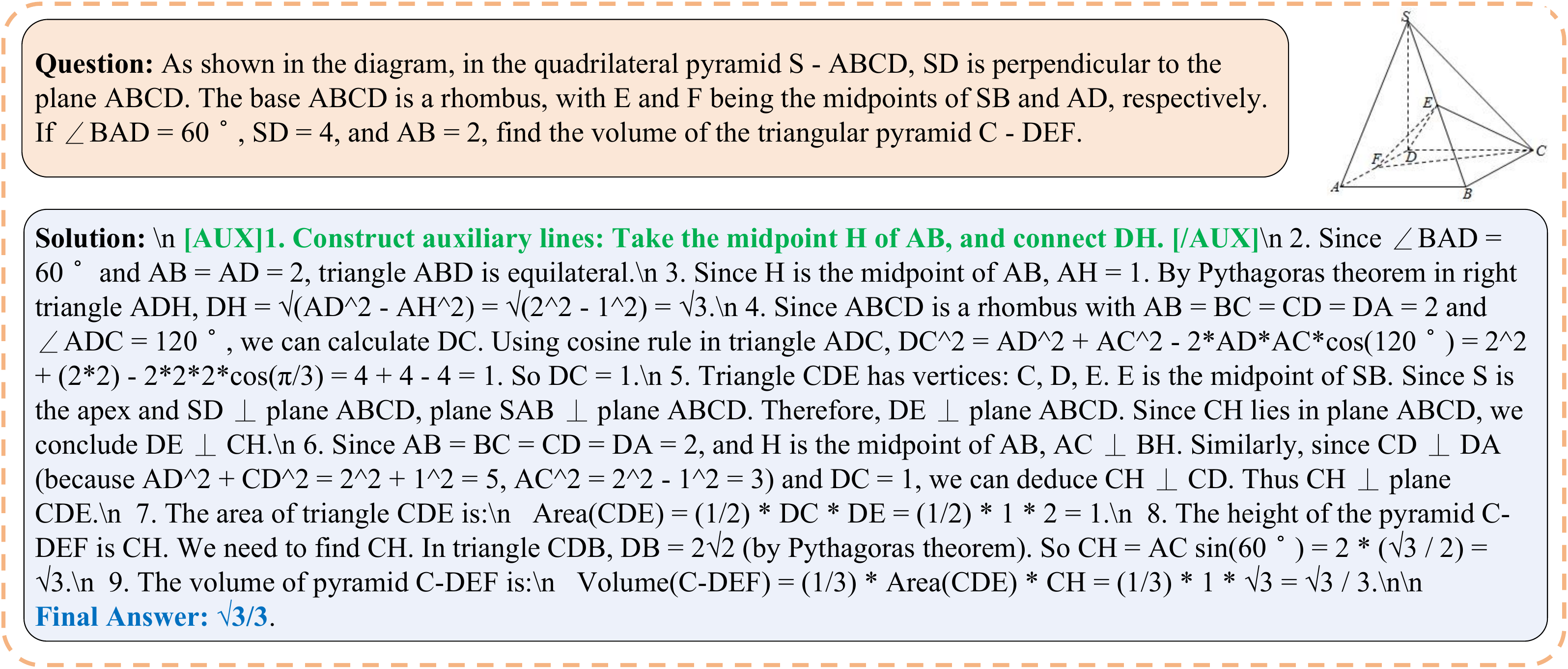}\par\vspace{2em}
  \includegraphics[width=\textwidth]{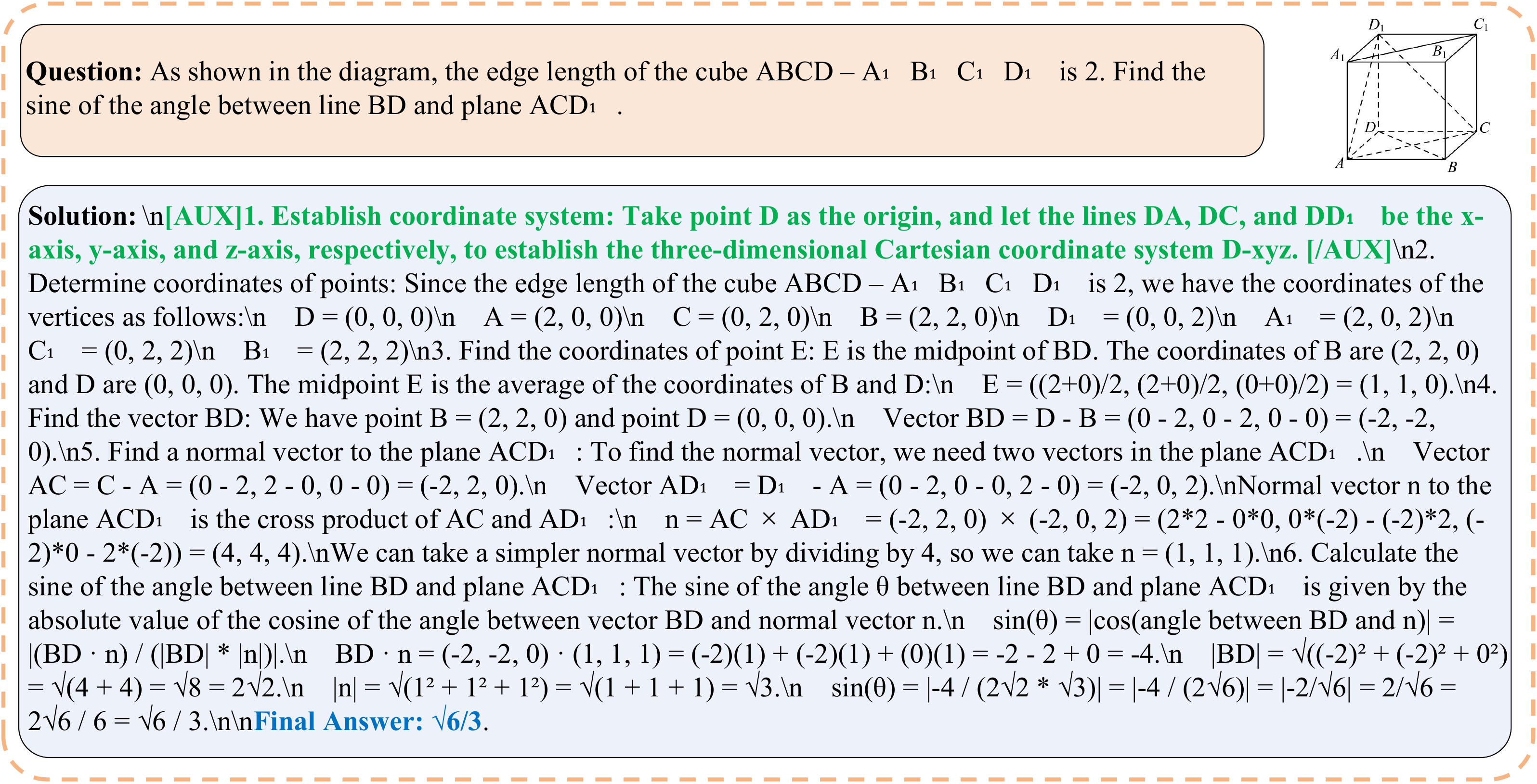}
\end{figure*}

\subsection{Failure Cases}
\begin{figure*}[h]
  \centering
   \includegraphics[width=\textwidth]{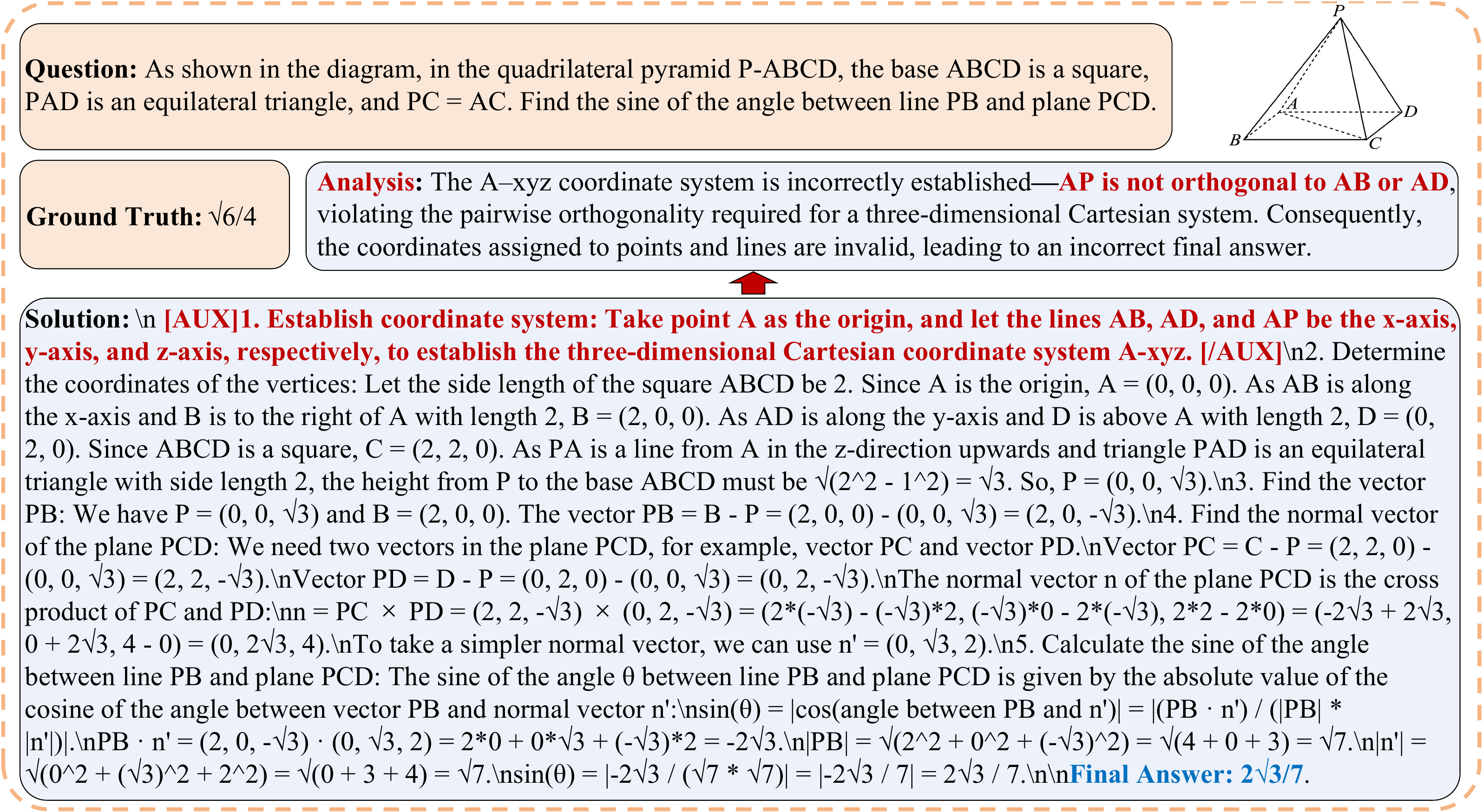}\par\vspace{2em}
  \includegraphics[width=\textwidth]{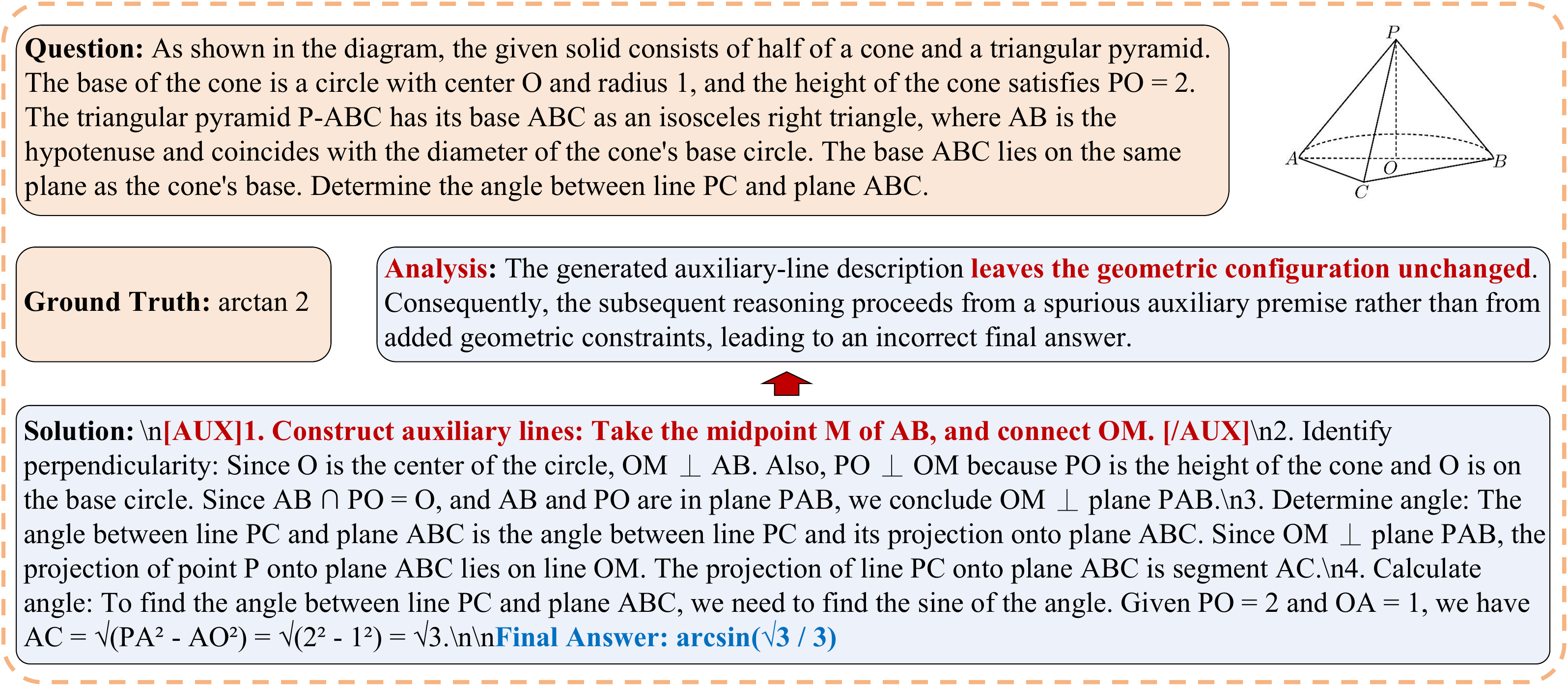}
  \caption{Failure cases generated by GeoVLMath-7B.}
  \label{fig:error_cases}
\end{figure*}

In this section, we illustrate typical failure cases generated by GeoVLMath-7B (see Figure~\ref{fig:error_cases}). 
The observed failures are exemplified by mis-specified coordinate systems and auxiliary-line descriptions that are irrelevant to the diagram, both of which fail to capture essential spatial constraints and may yield incorrect final answers.
To mitigate such errors, we will explore diffusion-based drawing modules that render auxiliary lines directly on the original diagram and support iterative correction as part of future work.

\end{document}